\PassOptionsToPackage{table}{xcolor}
\documentclass{selfevolagent}

\usepackage{amsmath}
\usepackage{amssymb}
\usepackage{colortbl}
\usepackage{float}
\usepackage{subcaption}
\usepackage{wrapfig}
\usepackage{needspace}
\usepackage{fontawesome5}
\usepackage{tabularx}
\usepackage{listings}

\newcommand{\corrauth}{\text{\raisebox{-0.12ex}{\scalebox{0.78}{\faEnvelope}}}}

\floatstyle{ruled}
\newfloat{algorithm}{tbp}{loa}
\floatname{algorithm}{Algorithm}

\graphicspath{{./}{figures/}}

\newcommand{\method}{Harness-R1}
\newcommand{\targetagent}{target agent}
\newcommand{\engineer}{harness engineer}

\newcommand{\paramtable}[1]{%
  \begin{center}
  \small
  \begin{tabularx}{0.78\linewidth}{@{}>{\raggedright\arraybackslash}X
    >{\raggedleft\arraybackslash}p{0.32\linewidth}@{}}
  \toprule
  Parameter & Value \\
  \midrule
  #1
  \bottomrule
  \end{tabularx}
  \end{center}
}

\definecolor{promptheader}{HTML}{B9D8E8}
\definecolor{promptframe}{HTML}{303030}
\newtcblisting{promptbox}[1]{%
  enhanced,
  breakable,
  listing only,
  listing engine=listings,
  colback=white,
  colframe=promptframe,
  colbacktitle=promptheader,
  coltitle=black,
  title={#1},
  fonttitle=\bfseries\small,
  center title,
  boxrule=0.45pt,
  arc=0pt,
  outer arc=0pt,
  left=1.6mm,
  right=1.6mm,
  top=1.0mm,
  bottom=1.0mm,
  listing options={%
    basicstyle=\ttfamily\fontsize{8.2}{9.6}\selectfont,
    breaklines=true,
    breakatwhitespace=false,
    columns=fullflexible,
    keepspaces=true,
    showstringspaces=false,
    upquote=true
  }
}

\title{Harness-R1: Learning to Edit Executable Runtime Harnesses from Agent Failure Trajectories}

\author[1,2\ddagger*]{Shuai Shao}
\author[1,2\ddagger*]{~Kangning Zhang}
\author[1,2*]{~Qingyao Li}
\author[3]{~Shijian Wang}
\author[2]{~Hao Wang}
\author[2,\corrauth]{\\Wenxiang Jiao}
\author[2,\corrauth]{~Yuan Lu}
\author[2]{~Yi Guo}
\author[1,\corrauth]{~Weiwen Liu}
\author[1,\corrauth]{~Weinan Zhang}

\affiliation[1]{Shanghai Jiao Tong University}
\affiliation[2]{Xiaohongshu Inc.}
\affiliation[3]{Southeast University}

\contribution[*]{Work done during internship at Xiaohongshu Inc.}
\contribution[\ddagger]{Equal contribution}
\contribution[\corrauth]{Corresponding authors}

\metadata[\raisebox{-0.18ex}{\scalebox{0.89}{\faEnvelope}}~Contact]{shaoshuai.ederson@sjtu.edu.cn, wenxiangjiaonju@gmail.com, liuww@sjtu.edu.cn}
\metadata[\raisebox{-0.14ex}{\scalebox{0.92}{\faGithub}}~Code]{\url{https://github.com/DeepExperience/Harness-R1}}
\metadata[\raisebox{-0.14ex}{\scalebox{0.92}{\faCube}}~Models]{\url{https://huggingface.co/ShaoShuai0605/Harness-R1}}

\abstract{
Agents built around large language models continually accumulate interaction trajectories during deployment, yet their behavior typically remains fixed.
Beyond updating model weights, these trajectories can improve the agent harness that constructs context, mediates tools, validates actions, and recovers execution.
We introduce \method{}, the first method, to our knowledge, that makes failure-conditioned, lifecycle-wide editing of an existing executable runtime a learned capability. It post-trains a dedicated harness engineer with online reinforcement learning so that its edits are optimized for the realized task success they produce, rather than proposed by a fixed editor.
A separate 9B engineer converts batches of target-agent failures into validated executable patches; fresh same-batch reruns of the frozen target provide outcome rewards, so training updates only the engineer.
Cold-start supervised fine-tuning initializes this editing policy, which is then trained online with group-relative policy optimization.
Across WebShop, ALFWorld, and DBBench, \method{} raises vanilla Qwen3.5-9B success from 44.3\% to 53.6\% (\(+9.3\) percentage points).
After direct target-agent fine-tuning, a target-specific engineer raises the average further from 59.2\% to 64.2\% (\(+5.0\) points); because these gains hold both before and after fine-tuning the target, \method{} points toward co-evolving the harness engineer and the target agent.
}

\begin{document}

\maketitle

\section{Introduction}

Large language models serve as the decision core of tool-using agents, enabling them to interpret tasks, maintain state, and pursue complex goals through multi-turn interaction with external environments \citep{yao2023react,liang2024encouraging,li2026deepagent,zhang2026looptool}.
Unlike a single model invocation, a deployed agent continually produces trajectories containing observations, actions, environment feedback, and task outcomes.
These trajectories record successful experience, but they also expose systematic failures such as tool misuse, lost state, protocol violations, repeated attempts, and failed recovery.
This raises a natural question: can agents use their interaction experience to improve continually rather than remain fixed after deployment?
This experience-to-improvement loop is a central concern of self-evolving agents \citep{gao2026selfevolvingagents,wu2026evolver,yu2026selfconsolidation}.

An agent system can improve at two complementary locations.
One line updates model parameters through supervised fine-tuning, reinforcement learning, or online learning, directly improving the actor that makes task decisions \citep{xia2026skillrl,lu2026skill0,shi2026skill1}.
The other keeps the model fixed and optimizes the \emph{agent harness} around it.
Context construction, memory and skills, tool mediation, action validation, and control and recovery logic are all harness components \citep{shinn2023reflexion,zhao2024expel,karten2026continualharness}.
Together, they determine what the model observes, which actions it can execute, how it interprets environment feedback, and how execution recovers after deviations.
Identical model weights can therefore yield substantially different agent capabilities under different harnesses.
Harness optimization offers a complementary path to model training: it improves the runtime mechanisms between a model and its environment without changing the model itself.

\begin{figure}[t]
    \centering
    \includegraphics[width=0.60\textwidth]{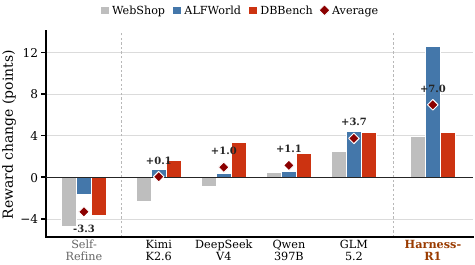}
    \caption{Matched-baseline reward changes across three benchmarks; diamonds denote the equal-weight average.}
    \label{fig:motivation-editing}
\end{figure}

Direct harness modification is not uniformly reliable.
Figure~\ref{fig:motivation-editing} compares matched-baseline changes in mean environment reward across the three benchmarks.
A fixed Self-Refine rule \citep{madaan2023selfrefine} lowers reward on all three benchmarks, and the gains from frontier harness editors are unstable or limited, with some even reducing the WebShop reward.
Prompting strong but fixed models to edit the harness is therefore not reliable enough.

Beyond such prompted edits, recent systems build dedicated harness-optimization pipelines.
Meta-Harness, Agentic Harness Engineering, and AutoHarness use agentic proposers to jointly edit prompts, tools, memory, middleware, or control logic from harness state, execution traces, and task feedback; Life-Harness and HarnessX extend the editable surface to lifecycle interventions and typed components \citep{lee2026metaharness,lin2026ahe,lou2026autoharness,xu2026lifeharness,chen2026harnessx}.
Yet the harness proposer usually remains fixed: outcomes select or iteratively refine patches without directly updating proposer parameters; HarnessX uses cross-harness GRPO to train the task model, while AEGIS retains symbolic harness editing.
Complementary work optimizes prompts, demonstrations, memories, skills, or task-solving programs \citep{yang2024opro,khattab2024dspy,shinn2023reflexion,zhao2024expel,xia2026skillrl,lu2026skill0,shi2026skill1}, but typically isolates one artifact or constructs a new solution program.
A workflow may be part of a harness, but generating one differs from learning to install failure-conditioned executable interventions into an existing multi-stage runtime.
The latter must decide when to intervene from actual target-agent failures and coordinate context, state, action execution, and recovery.
This leaves a less studied question: can we post-train a dedicated harness engineer with online reinforcement learning, so that improving an existing executable runtime from observed failures becomes a learned capability?

Training a harness engineer directly poses two challenges.
First, the editable runtime spans interdependent execution stages, so unrestricted code edits can break existing interfaces, produce non-executable behavior, or introduce changes unrelated to task success.
Second, text form and static rules cannot determine patch quality; only the target agent's behavior after applying the patch can do so.
Training therefore requires a grounded feedback path from target-agent failures, through constrained executable edits, to the performance gain that updates the engineer.

We introduce \method{}, a training paradigm that post-trains a dedicated \engineer{} with online reinforcement learning while keeping the \targetagent{} frozen.
Conditioned on batches of target-agent failures, the engineer generates validated executable runtime patches; the patched target reruns the same tasks, and the realized performance change rewards only the engineer.
Cold-start supervised fine-tuning initializes the policy before online GRPO.
Across WebShop, ALFWorld, and DBBench, \method{} raises average success from 44.3\% to 53.6\% (\(+9.3\) points) for the vanilla target and from 59.2\% to 64.2\% (\(+5.0\) points) after direct target-agent fine-tuning.

The main contributions can be summarized as follows.

\begin{itemize}
    \item We formulate failure-conditioned, lifecycle-wide harness editing as an online reinforcement-learning problem for a dedicated engineer while keeping the target agent frozen.
    \item We develop \method{}, combining cold-start supervised fine-tuning with group-relative policy optimization over the realized utility of executable runtime patches, and it improves the vanilla target by 9.3 points across three interactive benchmarks.
    \item We show that the harness engineer and the target agent can co-evolve: after direct target-agent fine-tuning, a target-specific \method{} engineer adds a further 5.0 points.
\end{itemize}

\begin{figure*}[t]
    \centering
    \includegraphics[width=\textwidth,page=2,trim={89.7pt 195.9pt 298.1pt 62.0pt},clip]{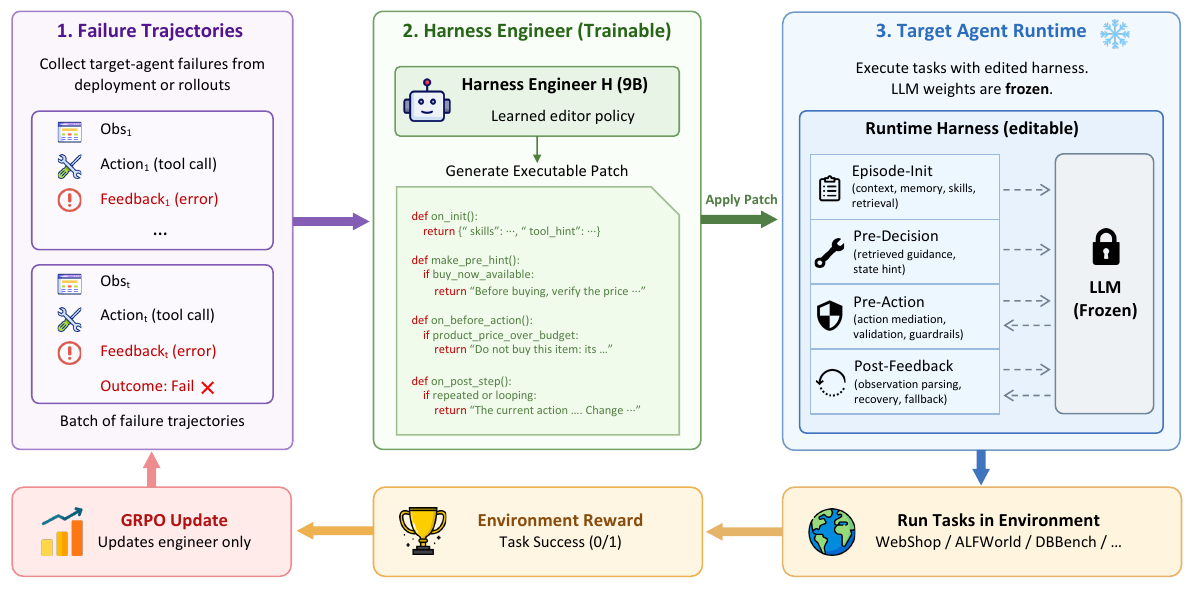}
    \caption{Overview of \method{}. Mined target failures become an evidence bundle (left); the harness engineer writes an executable patch that edits four lifecycle points in the runtime surrounding the frozen target (right); the patched target reruns the same tasks, and the resulting same-batch reward trains only the engineer via cold-start SFT and then GRPO (bottom loop). See the Method section for the loop, action space, and lifecycle hooks.}
    \label{fig:method-overview}
\end{figure*}

\section{Related Work}

\paragraph{LLM-Based Harness Evolution.}
Recent work treats the agent harness as an executable, multi-component optimization object.
One line searches or synthesizes whole harnesses from execution traces, evaluation scores, and rewards: Meta-Harness has a coding agent search over prior candidates, AutoHarness iteratively synthesizes code harnesses from environment-validity feedback, and AHE jointly evolves prompts, tools, middleware, skills, sub-agents, and memory \citep{lee2026metaharness,lou2026autoharness,lin2026ahe}.
A second line turns recurring interaction failures into scoped, regression-checked repairs across multiple execution stages, combining trace-grounded diagnosis with regression-aware validation \citep{xu2026lifeharness,chen2026harnessfix,zhang2026selfharness}.
As a closely related concurrent direction, HarnessX composes typed processors, performs symbolic trace-driven adaptation with AEGIS, and applies cross-harness GRPO to the task model \citep{chen2026harnessx}.
Across these systems the proposer may be a stronger external model or the target model itself, but none post-trains the proposer or editor weights from harness-editing outcomes; feedback instead guides program search, candidate selection, regression testing, or artifact promotion.
\method{} instead moves the learning target from the resulting harness to the editing policy: online reinforcement learning post-trains a dedicated harness engineer from the realized utility of its patches on a frozen target agent.

\paragraph{Algorithmic Optimization of Harness Components.}
A broader line optimizes prompts and other model-external artifacts within prespecified edit spaces and search procedures.
Some methods have an LLM propose and search over instruction candidates against a score, as in APE and OPRO \citep{zhou2023ape,yang2024opro}.
Others form natural-language ``gradients'' or textual feedback and propagate them to edit prompts or code, as in ProTeGi and TextGrad \citep{pryzant2023automatic,yuksekgonul2025textgrad}.
Population-based methods such as EvoPrompt, Promptbreeder, and GEPA mutate and select prompts by fitness \citep{guo2024evoprompt,fernando2024promptbreeder,agrawal2026gepa}, whereas pipeline compilers such as DSPy and MIPRO separate program structure from module parameters and search over instructions and bootstrapped demonstrations \citep{khattab2024dspy,opsahlong2024mipro}.
These methods can invoke strong language models and maintain histories, populations, or learned surrogates, but they generally do not post-train the proposer, reflector, or backward engine from editing outcomes; even when some fine-tune task modules, the product is a task-specific artifact or parameter rather than a dedicated harness-editor policy trained by editing outcomes.
\method{} instead trains the editor from the realized execution effects of its patches and modifies multiple stages of an existing target-agent runtime.

\paragraph{Learned Harness Editors.}
More directly related work post-trains policies that edit or control model-external structures, but each either restricts the edit space or couples editing with task solving.
Some train a dedicated editor from downstream outcomes over a narrow target, such as an independent context field or a revisable skill bank \citep{chen2026learningtoselfevolve,vishe2026skillr1,li2026codeskill}, or learn to generate task-solving workflows run by a frozen executor \citep{li2024autoflow,nie2025weakforstrong,zhang2026flowsteer}.
Closer to the runtime harness, others instruction-tune observation and action projections without reinforcement learning or persistent code patches, select among a few predefined structural actions under offline RL, or fold harness edits into a single task actor's action space \citep{wang2026harnessbridge,yi2026offlineharnesscontrol,luo2026hase}.
In contrast, \method{} isolates harness editing as a learning problem in its own right, casting failure-conditioned, lifecycle-wide editing as a standalone online reinforcement-learning task for a dedicated engineer, trained from the realized task outcomes of its executable patches while the target agent stays frozen.

\section{Method}
\label{sec:method}

In this section, we introduce \method{}, an online, outcome-grounded framework that post-trains a dedicated engineer to improve the executable runtime surrounding a frozen target agent.
We first formulate harness editing as a batch-conditioned learning problem in which each modification is evaluated by rerunning the same tasks.
We then describe where the modification can intervene across the agent lifecycle and how cold-start supervised fine-tuning followed by GRPO learns the editing policy.
Figure~\ref{fig:method-overview} summarizes this failure-to-edit-to-rerun loop and the separation between the trainable engineer and the frozen task agent.

\subsection{Problem Setup}

Let \(A\) denote a frozen target agent (its model together with the surrounding base runtime), and let \(B=\{x_i\}_{i=1}^{n}\) be a batch of \(n\) tasks in environment \(E\).
Within an episode the agent interacts with the environment over multiple turns: at each turn it reads the accumulated history and the current observation, and its frozen policy proposes an action.
Actions are expressed in the environment's native interface: structured tool or API calls and textual commands, such as \texttt{search} and \texttt{click} in WebShop, navigation and object manipulation in ALFWorld, and SQL queries in DBBench.
The environment executes the action, returns the next observation, and emits an outcome reward once the episode ends.
The component we adapt is the base runtime: the code that assembles the context shown to the model, forwards each action to the environment, and relays the feedback back to the agent.
This surrounding runtime, and not the model's weights, is exactly what the harness edits.
Running the unmodified agent over the batch yields baseline trajectories \(\tau_i^0\) and rewards \(R_i^0\).
A deterministic extractor retains only failed episodes and compacts their task constraints, selected action--observation excerpts, outcomes, and necessary environment state into a failure packet \(s_B\).
The engineer \(H_\theta\) reads this packet once and generates a batch-conditioned executable overlay \(P\); it neither answers the tasks nor participates in their rollouts.

The overlay \(P\) wraps this loop as executable hooks at four lifecycle points, leaving the agent's weights untouched (Figure~\ref{fig:method-overview}, right): (i)~\emph{episode initialization} sets up the starting context and episode state; (ii)~\emph{pre-decision} augments the context with retrieved guidance and interface constraints before the agent decides; (iii)~\emph{pre-action} is a runtime guardrail that may canonicalize, rewrite, or veto the proposed action before it reaches the environment; and (iv)~\emph{post-feedback} inspects the returned observation and triggers recovery when the trajectory stalls.
These hooks thus touch only the inputs and outputs surrounding the frozen policy, never the policy itself, and patches are validated before installation, with invalid patches having no effect.
Appendix~\ref{app:patch-interface} gives the invocation point and the permitted return effect of each hook.

After validation, the overlay is installed and the same frozen target reruns every task in \(B\), including tasks that originally succeeded.
Let \(R_i^P\) denote the resulting reward.
Define the full-batch performance difference and the engineer reward as
\begin{equation}
  \begin{aligned}
  \Delta_B(P)&=\frac{1}{n}\sum_{i=1}^{n}\left(R_i^P-R_i^0\right),\\
  r(B,P)&=
  \begin{cases}
    \Delta_B(P), & \text{if valid and complete},\\
    0, & \text{otherwise}.
  \end{cases}
  \end{aligned}
  \label{eq:patch-reward}
\end{equation}
Using the same tasks before and after editing controls task composition but defines a same-batch, transductive objective, with no iterative refinement within an instance or persistent patch memory across batches.
This reward is non-differentiable and observable only after an edit changes target-agent behavior, so it cannot be optimized directly.

\subsection{Outcome-Grounded Post-Training}
\method{} learns the editing policy in two stages.
Cold-start supervised fine-tuning first initializes a prior over valid, executable edits, and online, outcome-grounded GRPO then optimizes the realized task utility of patches applied to the frozen target.

\paragraph{Cold-start supervised fine-tuning.}
We first run the frozen target with its base harness and form editing instances from the resulting failed trajectories.
The teacher and RL instances use disjoint task batches.
A strong teacher proposes serialized editing responses \(y_j^T\) from the compact failure packets \(s_j\); we validate and evaluate their parsed overlays with the frozen target, retaining at most one executable, complete, non-regressive response per packet.
The resulting dataset \(\mathcal D_{\mathrm{SFT}}=\{(s_j,y_j^T)\}_{j=1}^{M}\) initializes the engineer by teacher-forced next-token prediction:
\begin{equation}
  \begin{aligned}
  \mathcal L_{\mathrm{SFT}}(\theta)
  &=-\frac{1}{\sum_{j=1}^{M}|y_j^T|}
  \sum_{j=1}^{M}\sum_{t=1}^{|y_j^T|}\\[-2pt]
  &\quad \log H_\theta\!\left(y_{j,t}^T\mid s_j,y_{j,<t}^T\right).
  \end{aligned}\hspace{1.2em}
  \label{eq:sft-objective}
\end{equation}

\paragraph{Outcome-grounded GRPO.}
Starting from the supervised policy, we perform online GRPO \citep{shao2024deepseekmath} and sample \(K=8\) candidate patches from the current policy for each failure packet (Alg.~\ref{alg:training}, lines~4--5).
Each candidate is parsed and validated into a patch (line~6); each valid patch is then installed independently and evaluated by rerunning the frozen target on the same full task batch (line~7), while invalid, no-op, or incomplete evaluations receive zero reward under Eq.~\eqref{eq:patch-reward} (line~8).
For rewards \(r_k=r(B,P_k)\), let \(\mu_B\) and \(\sigma_B\) be the empirical mean and standard deviation within the eight candidates generated from the same packet, and normalize the rewards into advantages (line~10):
\begin{equation}
  \widehat A_k=\frac{r_k-\mu_B}{\sigma_B}.
  \label{eq:grpo-advantage}
\end{equation}
Let \(y_k=(y_{k,1},\ldots,y_{k,T_k})\) be the engineer response parsed into \(P_k\), and let
\(\rho_{k,t}(\theta)=H_\theta(y_{k,t}\mid s_B,y_{k,<t})/H_{\theta_{\mathrm{old}}}(y_{k,t}\mid s_B,y_{k,<t})\).
The sequence-level advantage is shared by all response tokens, and the engineer maximizes the token-averaged clipped surrogate
\begin{equation}
  \begin{aligned}
  g_{k,t}(\theta)&=\min\!\Big\{
    \rho_{k,t}\widehat A_k,\\[-2pt]
    &\quad \operatorname{clip}(\rho_{k,t},1-\epsilon_\ell,1+\epsilon_h)
    \widehat A_k\Big\},\\
  \mathcal J(\theta)&=\mathbb E\!\left[
    \frac{1}{K}\sum_{k=1}^{K}\frac{1}{T_k}
    \sum_{t=1}^{T_k}w_{k,t}g_{k,t}(\theta)\right].
  \end{aligned}
  \label{eq:grpo-objective}
\end{equation}
Here \(\ell^{\mathrm{tr}}_{k,t}\) and \(\ell^{\mathrm{ro}}_{k,t}\) are the old-policy token log probabilities recomputed by the training engine and recorded by the rollout engine, respectively; \(w_{k,t}=\operatorname{clip}(\exp(\ell^{\mathrm{tr}}_{k,t}-\ell^{\mathrm{ro}}_{k,t}),0,2)\) is the truncated importance weight.
WebShop supplies shaped environment reward, whereas ALFWorld and DBBench supply binary success; no format-validity bonus or explicit KL loss is added.
Only the engineer parameters \(\theta\) are updated (line~12), and the outer loop iterates over update bundles until the training budget is exhausted (lines~2 and~13).

Algorithm~\ref{alg:training} summarizes the online RL stage.
Base trajectories, rewards, and failure packets are cached before optimization; online evaluation reruns only the patched target for candidates sampled from the current engineer.

\begin{algorithm}[t]
\caption{Online RL for harness editing.}
\label{alg:training}
\small
\textbf{Input:} Frozen target \(A\), cached RL records
\(\mathcal Q_{\mathrm{RL}}=\{(B,s_B,\mathbf R_B^0)\}\), initialized engineer
\(H_\theta\)\\
\textbf{Parameters:} group size \(K\), clipping bounds \(\epsilon_\ell,\epsilon_h\), learning rate \(\eta\), update-bundle size \(|\mathcal U|\), training budget\\
\textbf{Output:} Trained harness engineer \(H_{\theta^\star}\)
\vspace{2pt}

\begin{tabular}{@{}r@{\hspace{0.45em}}p{\dimexpr\columnwidth-2.2em\relax}@{}}
1: & \textbf{repeat} \\
2: & \quad Sample an update bundle \(\mathcal U\subset\mathcal Q_{\mathrm{RL}}\);
      set \(\theta_{\mathrm{old}}\leftarrow\theta\). \\
3: & \quad \textbf{for each} \((B,s_B,\mathbf R_B^0)\in\mathcal U\) \textbf{do} \\
4: & \qquad Sample \(\{y_k\}_{k=1}^{K}\sim
      H_{\theta_{\mathrm{old}}}(\cdot\mid s_B)\). \\
5: & \qquad \textbf{for} \(k=1,\ldots,K\) \textbf{do} \\
6: & \qquad\quad Parse and validate \(y_k\) into patch \(P_k\). \\
7: & \qquad\quad Independently install valid \(P_k\), rerun frozen \(A\)
      on all tasks in \(B\), and compute \(r_k\) with Eq.~\eqref{eq:patch-reward}. \\
8: & \qquad\quad Set \(r_k=0\) for invalid, no-op, or ultimately incomplete evaluations. \\
9: & \qquad \textbf{end for} \\
10: & \qquad Normalize \(\{r_k\}_{k=1}^{K}\) into \(\{\widehat A_k\}\)
       with Eq.~\eqref{eq:grpo-advantage}. \\
11: & \quad \textbf{end for} \\
12: & \quad Update only \(\theta\) with Eq.~\eqref{eq:grpo-objective}. \\
13: & \textbf{until} the training budget is exhausted; \textbf{return} \(H_\theta\). \\
\end{tabular}
\end{algorithm}
\FloatBarrier

\section{Experiments}

We evaluate \method{} on three interactive environments that stress different forms of agent execution.
We ask whether outcome-trained harness editing improves over supervised editing and strong fixed editors, remains useful after direct target-agent training, transfers to unseen target models and tasks, and which lifecycle positions drive the gains.

\begin{table*}[t]
\centering
\small
\setlength{\tabcolsep}{6.40pt}
\renewcommand{\arraystretch}{1.04}
\resizebox{\textwidth}{!}{%
\begin{tabular}{lccccccccccc}
\toprule
\textbf{Method} & \multicolumn{7}{c}{\textbf{ALFWorld}} & \multicolumn{2}{c}{\textbf{WebShop}} & \textbf{DBBench} & \textbf{Avg.} \\
\cmidrule(lr){2-8}\cmidrule(lr){9-10}\cmidrule(lr){11-11}\cmidrule(l){12-12}
 & Pick & Look & Clean & Heat & Cool & Pick2 & All & Score & Succ. & Succ. & \\
\midrule
\textcolor{black!55}{Qwen3.5-9B (with default harness)} & \textcolor{black!55}{75.8} & \textcolor{black!55}{66.7} & \textcolor{black!55}{16.9} & \textcolor{black!55}{28.8} & \textcolor{black!55}{9.2} & \textcolor{black!55}{47.5} & \textcolor{black!55}{40.6} & \textcolor{black!55}{66.0} & \textcolor{black!55}{31.2} & \textcolor{black!55}{61.0} & \textcolor{black!55}{44.3} \\
\addlinespace[1.5pt]
\rowcolor{black!8} \multicolumn{12}{@{}l}{\textbf{Prompt-based Agentic Methods}} \\
ReAct & 79.8 & 54.8 & 27.3 & 27.4 & 10.3 & 53.3 & 43.4 & 66.8 & 37.4 & 61.7 & 47.5 \\
Self-Refine & 70.7 & 45.2 & 11.7 & 28.8 & 6.9 & 57.4 & 39.0 & 61.7 & 29.0 & 57.3 & 41.8 \\
Reflection & 91.9 & 90.5 & 24.7 & 56.2 & 19.5 & 73.8 & 59.2 & 61.2 & 43.6 & 64.7 & 55.8 \\
\addlinespace[1.5pt]
\rowcolor{black!8} \multicolumn{12}{@{}l}{\textbf{Frontier Models}} \\
Qwen3.5-397B & 76.8 & 66.7 & 14.3 & 31.5 & 11.5 & 47.5 & 41.2 & 66.4 & 32.8 & 63.3 & 45.8 \\
GLM-5.2 & 77.8 & 78.6 & 18.2 & 27.4 & 16.1 & 54.9 & 45.0 & 68.4 & 36.0 & \cellcolor{blue!10}{65.3} & 48.8 \\
Kimi-K2.6 & 73.7 & 71.4 & 14.3 & 30.1 & 12.6 & 49.2 & 41.4 & 63.7 & 31.8 & 62.7 & 45.3 \\
DeepSeek-V4-Pro & 72.7 & 69.0 & 13.0 & 30.1 & 16.1 & 47.5 & 41.0 & 65.1 & 32.4 & 64.3 & 45.9 \\
Gemini-3.5-Flash & 67.7 & 69.0 & 14.3 & 24.7 & 10.3 & 35.2 & 35.4 & 63.2 & 33.6 & 64.0 & 44.3 \\
GPT-5.5 & 80.8 & 78.6 & 31.2 & 28.8 & 17.2 & 35.2 & 43.2 & 61.4 & 36.6 & 64.0 & 47.9 \\
\addlinespace[1.5pt]
\rowcolor{black!8} \multicolumn{12}{@{}l}{\textbf{Ours}} \\
Supervised-only engineer & 80.8 & 66.7 & 22.1 & 24.7 & 11.5 & 36.1 & 39.4 & 67.8 & 38.6 & 61.3 & 46.4 \\
\textbf{Harness-R1} & 77.8 & \cellcolor{blue!10}{81.0} & \cellcolor{blue!10}{58.4} & 43.8 & 34.5 & 39.3 & 53.2 & \cellcolor{blue!10}{69.9} & 42.2 & \cellcolor{blue!10}{65.3} & 53.6 \\
Agent SFT & \cellcolor{blue!10}{91.9} & \cellcolor{red!14}\textbf{100.0} & 46.8 & \cellcolor{blue!10}{56.2} & \cellcolor{blue!10}{48.3} & \cellcolor{blue!10}{85.2} & \cellcolor{blue!10}{71.2} & \cellcolor{red!14}\textbf{71.5} & \cellcolor{blue!10}{42.6} & 63.7 & \cellcolor{blue!10}{59.2} \\
\textbf{Agent SFT + Harness-R1} & \cellcolor{red!14}\textbf{93.9} & \cellcolor{red!14}\textbf{100.0} & \cellcolor{red!14}\textbf{81.8} & \cellcolor{red!14}\textbf{74.0} & \cellcolor{red!14}\textbf{72.4} & \cellcolor{red!14}\textbf{86.1} & \cellcolor{red!14}\textbf{84.0} & 68.7 & \cellcolor{red!14}\textbf{43.0} & \cellcolor{red!14}\textbf{65.7} & \cellcolor{red!14}\textbf{64.2} \\
\bottomrule
\end{tabular}%
}
\caption{Main results across WebShop, ALFWorld, and DBBench (\%). Score is the mean shaped reward and Succ. is the task success rate. Avg. is the equal-weight average of WebShop Succ., ALFWorld All, and DBBench Succ. Across all non-Reflection rows, \colorbox{red!14}{red} and \colorbox{blue!10}{blue} mark the highest and second-highest distinct value in each column, respectively; ties share a color.}
\label{tab:main}
\end{table*}

\subsection{Experimental Setup}

\paragraph{Benchmarks.}
WebShop \citep{yao2022webshop} evaluates grounded web navigation: an agent must search, inspect, and purchase a product satisfying a natural-language request.
ALFWorld \citep{shridhar2021alfworld} is a text-based embodied environment whose household tasks require multi-step navigation, object manipulation, state tracking, and recovery.
DBBench from AgentBench \citep{liu2024agentbench} is a relational-database environment whose natural-language tasks require schema inspection, structured SQL querying, record manipulation, and result verification.
Together, the three environments expose complementary failures in long-horizon interaction, action execution, and interface compliance.

\paragraph{Target agents and comparisons.}
Our primary target is a frozen Qwen3.5-9B agent \citep{qwen2026qwen35}.
To test whether harness adaptation remains useful after improving the actor itself, we also evaluate the same backbone after direct task-agent SFT.
Beyond this primary target, we further probe cross-model transfer by applying the trained engineer to a broad set of target agents unseen during training.
We compare the unmodified target against four groups: fixed prompt-based agentic strategies (ReAct \citep{yao2023react}, Self-Refine \citep{madaan2023selfrefine}, and Reflection \citep{shinn2023reflexion}); strong frontier models prompted as harness engineers (Qwen3.5-397B, GLM-5.2, Kimi-K2.6, DeepSeek-V4-Pro, Gemini-3.5-Flash, and GPT-5.5) \citep{qwen2026qwen35,zai2026glm52,moonshotai2026kimik26,deepseekai2026deepseekv4highlyefficientmilliontoken,googledeepmind2026gemini35flash,openai2026gpt55}; a supervised-only engineer; and outcome-trained \method{}.
Within each benchmark, an editor is evaluated against the same target and task set without its generated patch.

\paragraph{Evaluation.}
We report task success on all three benchmarks and the shaped environment score on WebShop; success is computed over 500, 500, and 300 tasks for WebShop, ALFWorld, and DBBench, respectively.
ALFWorld additionally reports success across its six task families and a task-level micro-average (All), and we report the average across the three benchmarks (Avg.) as the overall summary.
Reflection is reported under a separate two-episode \(\mathrm{success@2}\) protocol: its success columns are cumulative over two episodes and its Score is measured after retrying first-episode failures, whereas all other rows report \(\mathrm{success@1}\).
It is therefore not ranked against single-episode methods.

\paragraph{Training and selection.}
The engineer is a separate 9B model initialized by cold-start SFT and then optimized with online GRPO while the target remains frozen.
The cold-start SFT set comprises roughly 1{,}000 executable editing examples proposed by a GPT-5.5 teacher and filtered by validation on the frozen target, and online GRPO trains on roughly 1{,}500 failure packets from a disjoint task split.
We select checkpoints using aggregate development performance and use the same executable patch interface for all trained variants.
Appendix~\ref{app:prompts} reproduces the engineer prompt template, Appendix~\ref{app:implementation} lists the training, decoding, and reward hyperparameters, and Appendix~\ref{app:data-splits} details the task-level SFT, RL, validation, and test splits.

\subsection{Main Results}

Table~\ref{tab:main} summarizes target-specific performance against prompt-based strategies, frontier engineers, and supervised engineer training.
We focus on task success as the primary metric.

\paragraph{Outcome-trained harness editing improves the frozen target.}
\method{} raises success on all three benchmarks and improves the equal-weight average from 44.3\% to 53.6\%, a gain of 9.3 percentage points.
The largest absolute gain is on ALFWorld, where success rises from 40.6\% to 53.2\%, while WebShop and DBBench also improve.
The outcome-trained engineer is 7.1 points above the supervised-only engineer.

\paragraph{A dedicated trained engineer is more effective than fixed alternatives.}
Among frontier engineers, the strongest is GLM-5.2 at a 48.8\% average, below \method{} at 53.6\%.
Prompt-based strategies are not uniformly beneficial: ReAct improves the average by 3.2 points, whereas Self-Refine reduces it by 2.5 points.
Reflection reaches 55.8\% cumulative success under its two-episode protocol, which is not directly comparable to the single-episode rows.

\paragraph{The harness engineer co-evolves with the target agent.}
Direct agent SFT raises the unmodified target to a 59.2\% average, and a target-specific \method{} engineer trained for this stronger actor raises it further to 64.2\%, an additional 5.0 points.
Harness editing therefore keeps improving the target even after the actor itself has been fine-tuned, showing that the engineer can co-evolve with the target agent rather than saturating once the agent improves.
The gain concentrates in task success, especially on ALFWorld; although a few individual metrics dip slightly, \method{} still improves the overall success of the fine-tuned agent.

\Needspace*{0.52\textheight}
\subsection{Generalization across Target Agents}

\begin{wrapfigure}[21]{r}{0.44\textwidth}
    \centering
    \vspace{-0.6\baselineskip}
    \includegraphics[width=0.42\textwidth]{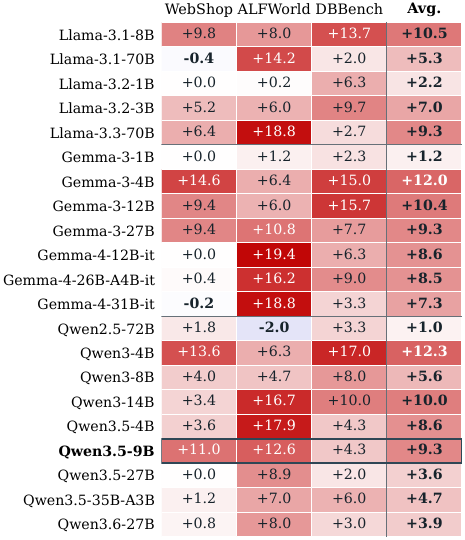}
    \caption{Target-agent generalization in success-rate points; Avg.\ weights benchmarks equally.}
    \label{fig:generalization}
\end{wrapfigure}

We next ask whether the learned editing policy can adapt to target models unseen during training.
Each target supplies its own failure traces and receives a newly generated patch, so this experiment tests editor-policy transfer rather than replaying a fixed patch.
Across twenty unseen target configurations, the benchmark-averaged gain is 7.06 percentage points, and every target-level average is positive.
Across the full 21-target matrix, 56 of 63 target--benchmark combinations improve, four are unchanged, and the three regressions are all small (\(\le 2.0\) points; Figure~\ref{fig:generalization}).
Aggregating matched tasks within each benchmark, gains stay positive at 4.15 points on WebShop, 9.63 on ALFWorld, and 7.37 on DBBench, with every delta computed on a matched target-specific task set.
The learned editing policy thus generalizes strongly: a single training recipe transfers to targets of different families and scales, improving every one without any per-target retuning.
Appendix~\ref{app:cross-target} reports the per-target success rates before and after patch installation that underlie these deltas.

\subsection{Held-Out Task Generalization}

We test whether sparse failures yield patches that improve unseen tasks.
For each benchmark and seed, every engineer observes the same 10 failures from the frozen Qwen3.5-9B target, generates one benchmark-specific patch, and applies it to all other tasks.
The pooled held-out set contains 1,270 tasks across WebShop, ALFWorld, and DBBench; we repeat the protocol over three matched seeds.

\begin{figure}[tb]
    \centering
    \begin{subfigure}[b]{0.49\textwidth}
        \centering
        \includegraphics[width=\linewidth]{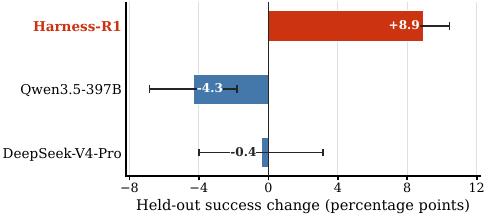}
        \caption{Held-out-task generalization from sparse failure evidence.}
        \label{fig:fresh-task-generalization}
    \end{subfigure}
    \hfill
    \begin{subfigure}[b]{0.49\textwidth}
        \centering
        \includegraphics[width=\linewidth]{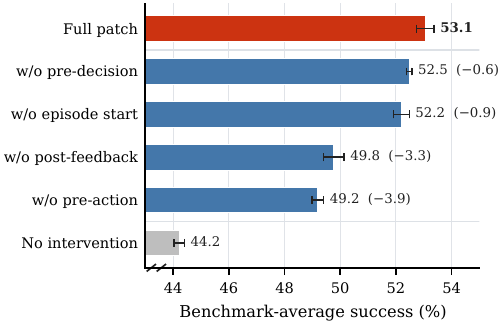}
        \caption{Fixed-patch lifecycle-position ablation.}
        \label{fig:lifecycle-ablation}
    \end{subfigure}
    \caption{Analysis of \method{} on the vanilla Qwen3.5-9B target. (a)~Held-out-task generalization from sparse failure evidence: bars show mean pooled success-rate change over three matched evidence seeds, and whiskers show sample standard deviation. (b)~Fixed-patch lifecycle-position ablation: success is benchmark-averaged and the horizontal axis is truncated.}
    \label{fig:analysis-combined}
\end{figure}

Figure~\ref{fig:fresh-task-generalization} shows that \method{} improves pooled held-out success by \(8.9\pm1.5\) percentage points and is positive for all three seeds.
Under the same protocol, Qwen3.5-397B and DeepSeek-V4-Pro yield \(-4.3\pm2.5\) and \(-0.4\pm3.6\) points, respectively.
The gap is not only in the mean: both frontier engineers straddle zero across seeds (spreads of \(\pm2.5\) and \(\pm3.6\) points around negative averages), swinging between marginal gains and sizable regressions, whereas \method{} stays positive on every seed at a tighter \(\pm1.5\).
Converting a handful of failures into a broadly useful edit is thus a capability that scale alone does not confer, and one that outcome-grounded training makes both stronger and more consistent.
Appendix~\ref{app:heldout} additionally reports how many seed-benchmark patches installed a real intervention and the corresponding full-split changes.

\subsection{Where in the Lifecycle Do Modifications Matter?}

Which intervention points account for the improvement?
Figure~\ref{fig:lifecycle-ablation} holds the frozen target and generated patches fixed, then disables one lifecycle position at a time alongside no-intervention and full-patch controls.
The no-intervention, full-patch, and leave-one-position-out arms each rerun the target three times per benchmark and use the same equal-benchmark weighting as Table~\ref{tab:main}.
Whiskers are standard deviations across benchmark recombinations.
The full patch reaches 53.1\% average success, 8.9 points above no intervention.
Removing pre-action mediation or post-feedback recovery reduces success by 3.9 and 3.3 points, whereas removing episode-start or pre-decision changes costs 0.9 and 0.6 points.
The dominant position depends on the environment: pre-action mediation matters most on WebShop, while post-feedback recovery matters most on ALFWorld.
Because a patch can coordinate multiple positions, and the evaluated WebShop patches contain only pre-action edits, these effects are conditional and should not be added into a universal importance ranking.
Appendix~\ref{app:lifecycle} tabulates the per-benchmark success rates behind this ablation.

\FloatBarrier

\section{Discussion}

As agents begin to improve other agents, harness editing becomes a form of AI improving AI.
In this setting, producing edits that merely look correct is not enough: an edit intervenes directly in a running executable system, so it must be precise, verifiable, and genuinely beneficial to the agent it modifies.
This is why \method{} learns from the realized task outcome of each patch rather than from whether its text appears reasonable.

\paragraph{Training a dedicated engineer beats prompting a larger model.}
As shown in the introduction, prompting strong but fixed frontier models to edit the harness is unreliable: they optimize for plausibility, emitting syntactically valid and reasonable-looking edits, but because they never rerun the target they cannot tell whether an edit actually raises task success, so their gains are unstable and sometimes even lower reward.
\method{} instead trains on the realized rerun outcome of each patch and learns edits that are genuinely useful rather than merely plausible: a valid, well-formed patch is necessary but not sufficient, and what ultimately matters is whether rerunning the target confirms a task gain.
Because the signal comes from outcomes rather than model scale, a 9B engineer trained this way surpasses much larger frontier editors (GLM-5.2 at 48.8\% versus \method{} at 53.6\%).
Appendix~\ref{app:cases} inspects stored patches together with their runtime traces, including a frontier-editor patch whose plausible diagnosis compiles into behavior that lowers success.

\paragraph{Failure-conditioned learning beats fixed harness patterns.}
Fixed, hand-designed harness strategies such as Self-Refine, Reflection, and ReAct are often assumed to help agents in general, yet they apply one hand-crafted rule uniformly to every task, ignoring both the specific target's failure modes and whether a given intervention actually works.
Our results show this is not reliable: a fixed Self-Refine rule lowers reward on all three benchmarks (an average of 2.6 points), and ReAct yields only limited and inconsistent gains.
\method{} instead generates edits conditioned on the target agent's observed failures and keeps only patches verified to help by rerun, so its improvements adapt to each target rather than betting on a single universal recipe.
This adaptivity also appears across the lifecycle: the dominant intervention point varies by environment (pre-action mediation on WebShop, post-feedback recovery on ALFWorld), which a fixed strategy cannot select on its own.

\paragraph{Limitations and future work.}
In this work, we study a single adaptation from a vanilla target to a fine-tuned one, where a re-trained engineer still improves the stronger actor.
A natural extension is to iterate this into multi-round co-evolution that alternates updates to the target agent and the harness engineer, so that gains in one continually reshape the training signal for the other; how such alternation converges and whether it yields compounding gains is a promising path toward agents that keep improving after deployment.
Our reward is also computed from same-batch task outcomes, which keeps training grounded but ties the signal to the tasks used to mine failures.
Future work can enrich this reward with held-out performance, so that edits are explicitly optimized against regressions on unseen tasks, and with inference-efficiency terms, so that useful patches are not obtained at unnecessary runtime cost, letting a single engineer jointly balance utility, robustness, and cost.

\section{Conclusion}

In this paper, we formalize failure-conditioned, lifecycle-wide editing of an executable agent harness as an online reinforcement learning problem for a dedicated engineer, while keeping the target agent frozen.
We propose \method{}, which initializes this editing policy with cold-start supervised fine-tuning and then trains it online with GRPO, so that edits are optimized for the realized task utility of executable runtime patches rather than produced by a fixed editor.
Across WebShop, ALFWorld, and DBBench, \method{} improves every benchmark and raises the average success of the vanilla Qwen3.5-9B target from 44.3\% to 53.6\% (\(+9.3\) points), while an engineer retrained for a directly fine-tuned target further raises it from 59.2\% to 64.2\% (\(+5.0\) points).
The learned editor also generalizes: it transfers to twenty unseen target models with a positive gain on every one and improves 1{,}270 held-out tasks.
Together, these results show that harness construction is a learnable capability that complements weight updates and lets the engineer and target co-evolve.

\bibliographystyle{plainnat}
\bibliography{harness_r1}

@misc{yao2023react,
      title={ReAct: Synergizing Reasoning and Acting in Language Models}, 
      author={Shunyu Yao and Jeffrey Zhao and Dian Yu and Nan Du and Izhak Shafran and Karthik Narasimhan and Yuan Cao},
      year={2023},
      eprint={2210.03629},
      archivePrefix={arXiv},
      primaryClass={cs.CL},
      url={https://arxiv.org/abs/2210.03629}, 
}

@misc{shridhar2021alfworld,
      title={ALFWorld: Aligning Text and Embodied Environments for Interactive Learning}, 
      author={Mohit Shridhar and Xingdi Yuan and Marc-Alexandre Côté and Yonatan Bisk and Adam Trischler and Matthew Hausknecht},
      year={2021},
      eprint={2010.03768},
      archivePrefix={arXiv},
      primaryClass={cs.CL},
      url={https://arxiv.org/abs/2010.03768}, 
}

@misc{yao2022webshop,
      title={WebShop: Towards Scalable Real-World Web Interaction with Grounded Language Agents}, 
      author={Shunyu Yao and Howard Chen and John Yang and Karthik Narasimhan},
      year={2023},
      eprint={2207.01206},
      archivePrefix={arXiv},
      primaryClass={cs.CL},
      url={https://arxiv.org/abs/2207.01206}, 
}

@misc{liu2024agentbench,
      title={AgentBench: Evaluating LLMs as Agents}, 
      author={Xiao Liu and Hao Yu and Hanchen Zhang and Yifan Xu and Xuanyu Lei and Hanyu Lai and Yu Gu and Hangliang Ding and Kaiwen Men and Kejuan Yang and Shudan Zhang and Xiang Deng and Aohan Zeng and Zhengxiao Du and Chenhui Zhang and Sheng Shen and Tianjun Zhang and Yu Su and Huan Sun and Minlie Huang and Yuxiao Dong and Jie Tang},
      year={2025},
      eprint={2308.03688},
      archivePrefix={arXiv},
      primaryClass={cs.AI},
      url={https://arxiv.org/abs/2308.03688}, 
}

@misc{lin2026ahe,
      title={Agentic Harness Engineering: Observability-Driven Automatic Evolution of Coding-Agent Harnesses}, 
      author={Jiahang Lin and Shichun Liu and Chengjun Pan and Lizhi Lin and Shihan Dou and Zhiheng Xi and Xuanjing Huang and Hang Yan and Zhenhua Han and Tao Gui and Yu-Gang Jiang},
      year={2026},
      eprint={2604.25850},
      archivePrefix={arXiv},
      primaryClass={cs.CL},
      url={https://arxiv.org/abs/2604.25850}, 
}

@misc{lee2026metaharness,
      title={Meta-Harness: End-to-End Optimization of Model Harnesses}, 
      author={Yoonho Lee and Roshen Nair and Qizheng Zhang and Kangwook Lee and Omar Khattab and Chelsea Finn},
      year={2026},
      eprint={2603.28052},
      archivePrefix={arXiv},
      primaryClass={cs.AI},
      url={https://arxiv.org/abs/2603.28052}, 
}

@misc{lou2026autoharness,
      title={AutoHarness: improving LLM agents by automatically synthesizing a code harness}, 
      author={Xinghua Lou and Miguel Lázaro-Gredilla and Antoine Dedieu and Carter Wendelken and Wolfgang Lehrach and Kevin P. Murphy},
      year={2026},
      eprint={2603.03329},
      archivePrefix={arXiv},
      primaryClass={cs.CL},
      url={https://arxiv.org/abs/2603.03329}, 
}

@misc{xu2026lifeharness,
      title={Adapting the Interface, Not the Model: Runtime Harness Adaptation for Deterministic LLM Agents}, 
      author={Tianshi Xu and Huifeng Wen and Meng Li},
      year={2026},
      eprint={2605.22166},
      archivePrefix={arXiv},
      primaryClass={cs.AI},
      url={https://arxiv.org/abs/2605.22166}, 
}

@misc{karten2026continualharness,
      title={Continual Harness: Online Adaptation for Self-Improving Foundation Agents}, 
      author={Seth Karten and Joel Zhang and Tersoo Upaa Jr and Ruirong Feng and Wenzhe Li and Chengshuai Shi and Chi Jin and Kiran Vodrahalli},
      year={2026},
      eprint={2605.09998},
      archivePrefix={arXiv},
      primaryClass={cs.LG},
      url={https://arxiv.org/abs/2605.09998}, 
}

@misc{chen2026harnessx,
      title={HarnessX: A Composable, Adaptive, and Evolvable Agent Harness Foundry}, 
      author={Tingyang Chen and Shuo Lu and Kang Zhao and Weicheng Meng and Hanlin Teng and Tianhao Li and Chao Li and Xule Liu and Jian Liang and Zhizhong Zhang and Yuan Xie and Heng Qu and Kun Shao and Jian Luan},
      year={2026},
      eprint={2606.14249},
      archivePrefix={arXiv},
      primaryClass={cs.AI},
      url={https://arxiv.org/abs/2606.14249}, 
}

@misc{chen2026harnessfix,
      title={From Failed Trajectories to Reliable LLM Agents: Diagnosing and Repairing Harness Flaws}, 
      author={Mengzhuo Chen and Junjie Wang and Zhe Liu and Yawen Wang and Haiming Zheng and Qing Wang},
      year={2026},
      eprint={2606.06324},
      archivePrefix={arXiv},
      primaryClass={cs.SE},
      url={https://arxiv.org/abs/2606.06324}, 
}

@misc{zhang2026selfharness,
      title={Self-Harness: Harnesses That Improve Themselves}, 
      author={Hangfan Zhang and Shao Zhang and Kangcong Li and Chen Zhang and Yang Chen and Yiqun Zhang and Lei Bai and Shuyue Hu},
      year={2026},
      eprint={2606.09498},
      archivePrefix={arXiv},
      primaryClass={cs.CL},
      url={https://arxiv.org/abs/2606.09498}, 
}

@misc{shinn2023reflexion,
      title={Reflexion: Language Agents with Verbal Reinforcement Learning}, 
      author={Noah Shinn and Federico Cassano and Edward Berman and Ashwin Gopinath and Karthik Narasimhan and Shunyu Yao},
      year={2023},
      eprint={2303.11366},
      archivePrefix={arXiv},
      primaryClass={cs.AI},
      url={https://arxiv.org/abs/2303.11366}, 
}

@misc{madaan2023selfrefine,
      title={Self-Refine: Iterative Refinement with Self-Feedback}, 
      author={Aman Madaan and Niket Tandon and Prakhar Gupta and Skyler Hallinan and Luyu Gao and Sarah Wiegreffe and Uri Alon and Nouha Dziri and Shrimai Prabhumoye and Yiming Yang and Shashank Gupta and Bodhisattwa Prasad Majumder and Katherine Hermann and Sean Welleck and Amir Yazdanbakhsh and Peter Clark},
      year={2023},
      eprint={2303.17651},
      archivePrefix={arXiv},
      primaryClass={cs.CL},
      url={https://arxiv.org/abs/2303.17651}, 
}

@misc{zhao2024expel,
      title={ExpeL: LLM Agents Are Experiential Learners}, 
      author={Andrew Zhao and Daniel Huang and Quentin Xu and Matthieu Lin and Yong-Jin Liu and Gao Huang},
      year={2024},
      eprint={2308.10144},
      archivePrefix={arXiv},
      primaryClass={cs.LG},
      url={https://arxiv.org/abs/2308.10144}, 
}

@misc{xia2026skillrl,
      title={SkillRL: Evolving Agents via Recursive Skill-Augmented Reinforcement Learning}, 
      author={Peng Xia and Jianwen Chen and Hanyang Wang and Jiaqi Liu and Kaide Zeng and Yu Wang and Siwei Han and Yiyang Zhou and Xujiang Zhao and Haifeng Chen and Zeyu Zheng and Cihang Xie and Huaxiu Yao},
      year={2026},
      eprint={2602.08234},
      archivePrefix={arXiv},
      primaryClass={cs.LG},
      url={https://arxiv.org/abs/2602.08234}, 
}

@misc{lu2026skill0,
      title={SKILL0: In-Context Agentic Reinforcement Learning for Skill Internalization}, 
      author={Zhengxi Lu and Zhiyuan Yao and Jinyang Wu and Chengcheng Han and Qi Gu and Xunliang Cai and Weiming Lu and Jun Xiao and Yueting Zhuang and Yongliang Shen},
      year={2026},
      eprint={2604.02268},
      archivePrefix={arXiv},
      primaryClass={cs.LG},
      url={https://arxiv.org/abs/2604.02268}, 
}

@misc{shi2026skill1,
      title={Skill1: Unified Evolution of Skill-Augmented Agents via Reinforcement Learning}, 
      author={Yaorui Shi and Yuxin Chen and Zhengxi Lu and Yuchun Miao and Shugui Liu and Qi GU and Xunliang Cai and Xiang Wang and An Zhang},
      year={2026},
      eprint={2605.06130},
      archivePrefix={arXiv},
      primaryClass={cs.AI},
      url={https://arxiv.org/abs/2605.06130}, 
}

@misc{yang2024opro,
      title={Large Language Models as Optimizers}, 
      author={Chengrun Yang and Xuezhi Wang and Yifeng Lu and Hanxiao Liu and Quoc V. Le and Denny Zhou and Xinyun Chen},
      year={2024},
      eprint={2309.03409},
      archivePrefix={arXiv},
      primaryClass={cs.LG},
      url={https://arxiv.org/abs/2309.03409}, 
}

@misc{khattab2024dspy,
      title={DSPy: Compiling Declarative Language Model Calls into Self-Improving Pipelines}, 
      author={Omar Khattab and Arnav Singhvi and Paridhi Maheshwari and Zhiyuan Zhang and Keshav Santhanam and Sri Vardhamanan and Saiful Haq and Ashutosh Sharma and Thomas T. Joshi and Hanna Moazam and Heather Miller and Matei Zaharia and Christopher Potts},
      year={2023},
      eprint={2310.03714},
      archivePrefix={arXiv},
      primaryClass={cs.CL},
      url={https://arxiv.org/abs/2310.03714}, 
}

@article{yuksekgonul2025textgrad,
  title={Optimizing generative ai by backpropagating language model feedback},
  author={Yuksekgonul, Mert and Bianchi, Federico and Boen, Joseph and Liu, Sheng and Lu, Pan and Huang, Zhi and Guestrin, Carlos and Zou, James},
  journal={Nature},
  volume={639},
  number={8055},
  pages={609--616},
  year={2025},
  publisher={Nature Publishing Group UK London}
}

@misc{opsahlong2024mipro,
      title={Optimizing Instructions and Demonstrations for Multi-Stage Language Model Programs}, 
      author={Krista Opsahl-Ong and Michael J Ryan and Josh Purtell and David Broman and Christopher Potts and Matei Zaharia and Omar Khattab},
      year={2024},
      eprint={2406.11695},
      archivePrefix={arXiv},
      primaryClass={cs.CL},
      url={https://arxiv.org/abs/2406.11695}, 
}

@misc{agrawal2026gepa,
      title={GEPA: Reflective Prompt Evolution Can Outperform Reinforcement Learning}, 
      author={Lakshya A Agrawal and Shangyin Tan and Dilara Soylu and Noah Ziems and Rishi Khare and Krista Opsahl-Ong and Arnav Singhvi and Herumb Shandilya and Michael J Ryan and Meng Jiang and Christopher Potts and Koushik Sen and Alexandros G. Dimakis and Ion Stoica and Dan Klein and Matei Zaharia and Omar Khattab},
      year={2026},
      eprint={2507.19457},
      archivePrefix={arXiv},
      primaryClass={cs.CL},
      url={https://arxiv.org/abs/2507.19457}, 
}

@misc{zhou2023ape,
      title={Large Language Models Are Human-Level Prompt Engineers}, 
      author={Yongchao Zhou and Andrei Ioan Muresanu and Ziwen Han and Keiran Paster and Silviu Pitis and Harris Chan and Jimmy Ba},
      year={2023},
      eprint={2211.01910},
      archivePrefix={arXiv},
      primaryClass={cs.LG},
      url={https://arxiv.org/abs/2211.01910}, 
}

@misc{pryzant2023automatic,
      title={Automatic Prompt Optimization with "Gradient Descent" and Beam Search}, 
      author={Reid Pryzant and Dan Iter and Jerry Li and Yin Tat Lee and Chenguang Zhu and Michael Zeng},
      year={2023},
      eprint={2305.03495},
      archivePrefix={arXiv},
      primaryClass={cs.CL},
      url={https://arxiv.org/abs/2305.03495}, 
}

@misc{guo2024evoprompt,
      title={EvoPrompt: Connecting LLMs with Evolutionary Algorithms Yields Powerful Prompt Optimizers}, 
      author={Qingyan Guo and Rui Wang and Junliang Guo and Bei Li and Kaitao Song and Xu Tan and Guoqing Liu and Jiang Bian and Yujiu Yang},
      year={2025},
      eprint={2309.08532},
      archivePrefix={arXiv},
      primaryClass={cs.CL},
      url={https://arxiv.org/abs/2309.08532}, 
}

@misc{fernando2024promptbreeder,
      title={Promptbreeder: Self-Referential Self-Improvement Via Prompt Evolution}, 
      author={Chrisantha Fernando and Dylan Banarse and Henryk Michalewski and Simon Osindero and Tim Rocktäschel},
      year={2023},
      eprint={2309.16797},
      archivePrefix={arXiv},
      primaryClass={cs.CL},
      url={https://arxiv.org/abs/2309.16797}, 
}

@misc{chen2026learningtoselfevolve,
      title={Learning to Self-Evolve}, 
      author={Xiaoyin Chen and Canwen Xu and Yite Wang and Boyi Liu and Zhewei Yao and Yuxiong He},
      year={2026},
      eprint={2603.18620},
      archivePrefix={arXiv},
      primaryClass={cs.CL},
      url={https://arxiv.org/abs/2603.18620}, 
}

@misc{vishe2026skillr1,
      title={Skill-R1: Agent Skill Evolution via Reinforcement Learning}, 
      author={Yash Vishe and Rohan Surana and Xunyi Jiang and Zihan Huang and Xintong Li and Nikki Lijing Kuang and Tong Yu and Ryan A. Rossi and Jingbo Shang and Julian McAuley and Junda Wu},
      year={2026},
      eprint={2605.09359},
      archivePrefix={arXiv},
      primaryClass={cs.LG},
      url={https://arxiv.org/abs/2605.09359}, 
}

@misc{li2026codeskill,
      title={CODESKILL: Learning Self-Evolving Skills for Coding Agents}, 
      author={Yanzhou Li and Yiran Zhang and Xiaoyu Zhang and Xiaoxia Liu and Yang Liu},
      year={2026},
      eprint={2605.25430},
      archivePrefix={arXiv},
      primaryClass={cs.AI},
      url={https://arxiv.org/abs/2605.25430}, 
}

@misc{li2024autoflow,
      title={AutoFlow: Automated Workflow Generation for Large Language Model Agents}, 
      author={Zelong Li and Shuyuan Xu and Kai Mei and Wenyue Hua and Balaji Rama and Om Raheja and Hao Wang and He Zhu and Yongfeng Zhang},
      year={2024},
      eprint={2407.12821},
      archivePrefix={arXiv},
      primaryClass={cs.CL},
      url={https://arxiv.org/abs/2407.12821}, 
}

@misc{nie2025weakforstrong,
      title={Weak-for-Strong: Training Weak Meta-Agent to Harness Strong Executors}, 
      author={Fan Nie and Lan Feng and Haotian Ye and Weixin Liang and Pan Lu and Huaxiu Yao and Alexandre Alahi and James Zou},
      year={2025},
      eprint={2504.04785},
      archivePrefix={arXiv},
      primaryClass={cs.AI},
      url={https://arxiv.org/abs/2504.04785}, 
}

@misc{zhang2026flowsteer,
      title={FlowSteer: Towards Agents Designing Agentic Workflows via Reinforced Progressive Canvas Editing}, 
      author={Mingda Zhang and Wenjin Liu and Tiesunlong Shen and Qika Lin and Rui Mao and Erik Cambria and Xiaoying Tang and Haoran Luo},
      year={2026},
      eprint={2602.01664},
      archivePrefix={arXiv},
      primaryClass={cs.AI},
      url={https://arxiv.org/abs/2602.01664}, 
}

@misc{wang2026harnessbridge,
      title={HarnessBridge: Learnable Bidirectional Controller for LLM Agent Harness}, 
      author={Xiaoxuan Wang and Haixin Wang and Alexander Taylor and Jason Cong and Yizhou Sun and Wei Wang},
      year={2026},
      eprint={2606.12882},
      archivePrefix={arXiv},
      primaryClass={cs.AI},
      url={https://arxiv.org/abs/2606.12882}, 
}

@misc{yi2026offlineharnesscontrol,
      title={Learning to Control LLM Agent Harnesses with Offline Reinforcement Learning}, 
      author={Haiwen Yi and Xinyuan Song},
      year={2026},
      eprint={2607.05458},
      archivePrefix={arXiv},
      primaryClass={cs.LG},
      url={https://arxiv.org/abs/2607.05458}, 
}

@misc{luo2026hase,
      title={Harness-Aware Self-Evolving: Co-Evolving Model Weights, Harness, and Task Solutions}, 
      author={Haochen Luo and Yi Huang and Sichun Luo and Fengyuan Liu and Lei Li and Zefa Hu and Junlan Feng and Qi Liu},
      year={2026},
      eprint={2607.03935},
      archivePrefix={arXiv},
      primaryClass={cs.AI},
      url={https://arxiv.org/abs/2607.03935}, 
}

@misc{shao2024deepseekmath,
      title={DeepSeekMath: Pushing the Limits of Mathematical Reasoning in Open Language Models}, 
      author={Zhihong Shao and Peiyi Wang and Qihao Zhu and Runxin Xu and Junxiao Song and Xiao Bi and Haowei Zhang and Mingchuan Zhang and Y. K. Li and Y. Wu and Daya Guo},
      year={2024},
      eprint={2402.03300},
      archivePrefix={arXiv},
      primaryClass={cs.CL},
      url={https://arxiv.org/abs/2402.03300}, 
}

@misc{liang2024encouraging,
      title={Encouraging Divergent Thinking in Large Language Models through Multi-Agent Debate}, 
      author={Tian Liang and Zhiwei He and Wenxiang Jiao and Xing Wang and Yan Wang and Rui Wang and Yujiu Yang and Shuming Shi and Zhaopeng Tu},
      year={2024},
      eprint={2305.19118},
      archivePrefix={arXiv},
      primaryClass={cs.CL},
      url={https://arxiv.org/abs/2305.19118}, 
}

@misc{li2026deepagent,
      title={DeepAgent: A General Reasoning Agent with Scalable Toolsets}, 
      author={Xiaoxi Li and Wenxiang Jiao and Jiarui Jin and Guanting Dong and Jiajie Jin and Yinuo Wang and Hao Wang and Yutao Zhu and Ji-Rong Wen and Yuan Lu and Zhicheng Dou},
      year={2026},
      eprint={2510.21618},
      archivePrefix={arXiv},
      primaryClass={cs.AI},
      url={https://arxiv.org/abs/2510.21618}, 
}

@misc{zhang2026looptool,
      title={LoopTool: Closing the Data-Training Loop for Robust LLM Tool Calls}, 
      author={Kangning Zhang and Wenxiang Jiao and Kounianhua Du and Yuan Lu and Weiwen Liu and Weinan Zhang and Yong Yu},
      year={2025},
      eprint={2511.09148},
      archivePrefix={arXiv},
      primaryClass={cs.CL},
      url={https://arxiv.org/abs/2511.09148}, 
}

@article{gao2026selfevolvingagents,
  title         = {A Survey of Self-Evolving Agents: What, When, How, and Where to Evolve on the Path to Artificial Super Intelligence},
  author        = {Gao, Huan-ang and others},
  journal       = {Transactions on Machine Learning Research},
  year          = {2026},
  eprint        = {2507.21046},
  archivePrefix = {arXiv},
  primaryClass  = {cs.AI},
  doi           = {10.48550/arXiv.2507.21046},
  url           = {https://arxiv.org/abs/2507.21046}
}

@inproceedings{wu2026evolver,
  title         = {{EvolveR}: Self-Evolving {LLM} Agents through an Experience-Driven Lifecycle},
  author        = {Wu, Rong and Wang, Xiaoman and Mei, Jianbiao and Cai, Pinlong and Fu, Daocheng and Yang, Cheng and Wen, Licheng and Yang, Xuemeng and Shen, Yufan and Wang, Yuxin and Shi, Botian},
  booktitle     = {International Conference on Machine Learning},
  year          = {2026},
  eprint        = {2510.16079},
  archivePrefix = {arXiv},
  primaryClass  = {cs.CL},
  doi           = {10.48550/arXiv.2510.16079},
  url           = {https://arxiv.org/abs/2510.16079}
}

@misc{yu2026selfconsolidation,
  title         = {Self-Consolidation for Self-Evolving Agents},
  author        = {Yu, Hongzhuo and Zhu, Fei and Xie, Guo-Sen and Shao, Ling},
  year          = {2026},
  eprint        = {2602.01966},
  archivePrefix = {arXiv},
  primaryClass  = {cs.LG},
  doi           = {10.48550/arXiv.2602.01966},
  url           = {https://arxiv.org/abs/2602.01966}
}

@misc{zai2026glm52,
  author       = {{Z.ai}},
  title        = {{GLM-5.2}: Built for Long-Horizon Tasks},
  year         = {2026},
  month        = jun,
  howpublished = {\url{https://z.ai/blog/glm-5.2}},
  note         = {Accessed: 2026-07-28}
}

@misc{openai2026gpt55,
  author       = {{OpenAI}},
  title        = {Introducing {GPT-5.5}},
  year         = {2026},
  month        = apr,
  howpublished = {\url{https://openai.com/index/introducing-gpt-5-5/}},
  note         = {Accessed: 2026-07-28}
}

@misc{moonshotai2026kimik26,
  author       = {{Moonshot AI}},
  title        = {{Kimi K2.6}: Advancing Open-Source Coding},
  year         = {2026},
  month        = apr,
  howpublished = {\url{https://www.kimi.com/blog/kimi-k2-6}},
  note         = {Accessed: 2026-07-28}
}

@misc{qwen2026qwen35,
  author       = {{Qwen Team}},
  title        = {{Qwen3.5}: Towards Native Multimodal Agents},
  year         = {2026},
  month        = feb,
  howpublished = {\url{https://qwen.ai/blog?id=qwen3.5}},
  note         = {Accessed: 2026-07-28}
}

@misc{googledeepmind2026gemini35flash,
  author       = {{Google DeepMind}},
  title        = {{Gemini 3.5 Flash}: Model Card},
  year         = {2026},
  month        = may,
  day          = {19},
  howpublished = {\url{https://deepmind.google/models/model-cards/gemini-3-5-flash/}},
  note         = {Accessed: 2026-07-28}
}

@misc{deepseekai2026deepseekv4highlyefficientmilliontoken,
      title={DeepSeek-V4: Towards Highly Efficient Million-Token Context Intelligence}, 
      author={DeepSeek-AI and others},
      year={2026},
      eprint={2606.19348},
      archivePrefix={arXiv},
      primaryClass={cs.CL},
      url={https://arxiv.org/abs/2606.19348}, 
}

\clearpage
\beginappendix
\makeatletter
\@addtoreset{table}{section}
\@addtoreset{figure}{section}
\makeatother
\renewcommand{\thetable}{\thesection.\arabic{table}}
\renewcommand{\thefigure}{\thesection.\arabic{figure}}
\setcounter{table}{0}
\setcounter{figure}{0}


\section{Prompts}
\label{app:prompts}

This section presents the model-facing prompt template used to train and
evaluate the harness engineer.  Following the presentation style of prompt
appendices, fixed instructions are shown separately from the per-example
input.  Angle-bracketed strings denote substituted fields rather than literal
tokens.  SFT and online GRPO use the same response protocol; the latter
constructs the failure packet from the current frozen target-agent rollouts.
The released SFT JSON contains the complete materialized messages for all
three benchmarks.

\subsection{Prompt for the Harness Engineer}

\begin{promptbox}{System Instruction}
You are a Harness-R1 harness engineer. Given a batch of failed
<BENCHMARK> rollout traces, analyze recurring failures, then produce one
reusable <BENCHMARK> code-hook harness patch as a single JSON object inside
a <patch>...</patch> block.
\end{promptbox}

\begin{promptbox}{Input}
You will edit only the reusable <BENCHMARK> harness, not task answers.

The chat template has already opened the assistant thinking block. Continue
concise recurring-failure reasoning, close it with </think>, then output
exactly one <patch> block.

After your reasoning, output exactly one <patch> block containing a single
JSON patch object with top-level keys benchmark, description, and actions.
Output nothing after the </patch> block.

Patch JSON top-level contract:
- benchmark: exactly "<BENCHMARK_ID>"
- description: a short, general description
- actions: a non-empty array of ADD_CODE_HOOK objects

ADD_CODE_HOOK ::= {
  "type": "add_code_hook",
  "hook": "on_init" | "make_pre_hint" |
          "on_before_action" | "on_post_step",
  "code": "<PYTHON SOURCE DEFINING hook(ctx, nb)>"
}

Hook return contracts:
- on_init -> skills/tool_hint or None
- make_pre_hint -> message or None
- on_before_action -> block_and_prompt, rewrite_action, or force_action
- on_post_step -> inject_hint or, where supported, force_action

General rules:
- Infer reusable interventions from observable recurring failures.
- Do not encode task-specific answers, indices, or identifiers.
- Define hook(ctx, nb) without imports or global state.
- Put all JSON inside the <patch> block; output nothing after </patch>.

<BENCHMARK-SPECIFIC RUNTIME CONTEXT AND RESTRICTIONS>

Observed no-harness rollout evidence:

<FAILURE TRACE PACKET>
\end{promptbox}

\begin{center}
\begin{minipage}{0.92\linewidth}
\small
\textbf{Figure A.1:} System and input prompt template for the harness engineer.
The failure packet contains a group of failed trajectories from the frozen
target agent; the benchmark-specific insertion is reproduced below.
\end{minipage}
\end{center}

\subsection{Benchmark-Specific Prompt Content}

\begin{promptbox}{Benchmark-Specific Instruction}
[WebShop]
benchmark = "webshop"
Active action type: add_code_hook only.
Runtime context:
  action: tool, value, value_normalized, final_action
  state: page_type, clickables, remaining_steps, current_price, price_max,
         repeated-click/search and product-stall counters
  task: task_type, required_color, required_size, required_material
  webshop: search_queries, visited products, current product,
           selected attributes, available attribute options
  predicates: required_options_unselected, product_price_over_budget,
              repeated click/search, stalled product page,
              buy-now availability, action admissibility
Use pre-action mediation only for narrow, observable mistakes such as an
over-budget purchase, an unselected required option, or a repeated action.
Do not hard-code product IDs, titles, ASINs, task indices, or answers.

[ALFWorld]
benchmark = "alfworld"
Actions may contain two to four add_code_hook entries; use each hook at most
once and omit hooks not supported by recurring evidence.
Runtime context:
  action: raw, normalized, final_action, in_admissible
  state: repeated observation/action, invalid-action and remaining-step signals
  task: task_type, target_type, destination_type
  world: current location, inventory, object locations, visited locations,
         target facts, and placement facts
  admissible: actions currently accepted by the environment
Maintain reusable task-stage state from observations and completed actions.
Any exact mediated action must be selected from admissible actions. Do not
copy numbered object/location instances or complete task solutions.

[DBBench]
benchmark = "dbbench"
Actions may contain two to four add_code_hook entries; use each hook at most
once and omit hooks not supported by recurring evidence.
Runtime context:
  action: execute_sql or commit_final_answer, query, submitted answers
  state: SQL count/history, last result/error, error/empty/loop streaks,
         mutation status, candidate answer shape, remaining rounds
  task: task_type, answer_shape, target_table, description
  dbbench: SQL history, discovered columns, database response, round
  predicates: premature/empty commit, mutation not attempted, SQL error,
              empty result, unknown column, syntax error, repeated SQL,
              candidate answer available, remaining rounds low
DBBench hooks may inject guidance or block a pending action. They do not
rewrite SQL or force a commit. Do not hard-code cell values, exact answers,
task indices, or ground-truth SQL.
\end{promptbox}

\begin{center}
\begin{minipage}{0.92\linewidth}
\small
\textbf{Figure A.2:} Benchmark-specific content inserted into the engineer
prompt.  These fields expose runtime evidence rather than hidden task answers.
\end{minipage}
\end{center}

\subsection{Assistant Response Format}

\begin{promptbox}{Assistant Prefill and Target}
<think>
<concise analysis of recurring failures, the proposed reusable intervention,
and the regression risk>
</think>
<patch>
{
  "benchmark": "<BENCHMARK_ID>",
  "description": "<GENERAL PATCH DESCRIPTION>",
  "actions": [
    {
      "type": "add_code_hook",
      "hook": "<SELECTED LIFECYCLE POSITION>",
      "code": "def hook(ctx, nb):\n    ...\n    return <STRUCTURED EFFECT>"
    }
  ]
}
</patch>
\end{promptbox}

\begin{center}
\begin{minipage}{0.92\linewidth}
\small
\textbf{Figure A.3:} Harness-engineer response format.  The chat template
prefills the opening \texttt{<think>} token sequence; training targets contain
the remaining analysis and exactly one executable patch.
\end{minipage}
\end{center}

\section{Implementation Details}
\label{app:implementation}

Cold-start SFT, online GRPO, and direct target-agent SFT each run on a single
node with eight NVIDIA H800 GPUs.  The tables below list the settings needed to
reproduce each stage; framework defaults and settings that only affect memory
use, such as gradient checkpointing and the ZeRO stage, are omitted.

\subsection{Harness-Engineer Cold-Start SFT}

GPT-5.5 generates candidate patches from failure packets in the SFT task
split.  We retain candidates that are executable, complete the same-batch
rerun, and achieve a non-negative task-reward change.  This produces
approximately 1K teacher-filtered editing examples (877 in total: 381
WebShop, 248 ALFWorld, and 248 DBBench).

\paramtable{
Base model & Qwen3.5-9B \\
Training examples & 877 \\
Fine-tuning type & Full parameter \\
Epochs & 2 \\
Context length & 32,768 \\
Global batch size & 24 \\
Optimizer & AdamW \\
Learning rate & $10^{-5}$ \\
LR schedule & Cosine \\
Warmup ratio & 0.03 \\
Precision & BF16 \\
Seed & 42 \\
}

\subsection{Online GRPO}

\paramtable{
Candidates per prompt $K$ & 8 \\
Rollout batch size & 4 prompts \\
Global batch size & 32 sequences \\
Rollout iterations per update & 4 \\
Maximum policy staleness & 8 \\
Learning rate & $10^{-6}$ \\
LR schedule & Constant \\
Optimizer & Adam \\
Adam $\beta_1,\beta_2$ & 0.9, 0.98 \\
Weight decay & 0.1 \\
GRPO clip lower / upper & 0.20 / 0.28 \\
Truncated importance sampling & Enabled; maximum weight 2.0 \\
Entropy / explicit KL coefficient & 0 / 0 \\
Rollout temperature & 0.7 \\
Rollout top-$p$ & 0.95 \\
Maximum prompt length & 28,672 \\
Maximum response length & 12,288 \\
Seeds (training / rollout) & 1,234 / 42 \\
}

\subsection{Direct Target-Agent SFT}

For the sequential adaptation experiment, we directly fine-tune the
Qwen3.5-9B target agent on successful no-intervention trajectories from the
same task-level training split used to construct the engineer data.  We retain
one trajectory for each benchmark and canonical task identity, yielding 2,515
complete multi-turn episodes: 901 from WebShop, 774 from ALFWorld, and 840
from DBBench.  The optimization settings not listed below, including the
optimizer, learning rate, schedule, warmup, precision, and seed, are identical
to the cold-start SFT configuration above.

\paramtable{
Base model & Qwen3.5-9B \\
Training trajectories & 2,515 \\
WebShop / ALFWorld / DBBench & 901 / 774 / 840 \\
Trajectory selection & Successful and task-deduplicated \\
Epochs & 2 \\
Sequence cutoff length & 24,576 \\
Global batch size & 24 \\
Target thinking & Disabled \\
}

\subsection{Evaluation and Reward}

\paramtable{
Engineer reward & Full-batch mean reward change \(\Delta_B(P)\) \\
WebShop task reward & Native continuous environment reward \\
ALFWorld / DBBench task reward & Binary success \\
Invalid, no-op, or incomplete patch reward & 0 \\
Engineer temperature & 0 \\
Engineer maximum response length & 12,288 \\
Engineer thinking & Enabled, with \texttt{<think>} prefill \\
Target temperature & 0 \\
Target maximum response length & 4,096 \\
Target tool choice & \texttt{auto} \\
Target thinking & Disabled \\
WebShop goal seed & 233 \\
}

\section{Executable Patch Interface}
\label{app:patch-interface}

A patch can intervene at four positions in the target agent's execution
lifecycle.  Each hook receives benchmark-specific runtime context and returns
only an effect defined by the host runtime.

\begin{table}[h]
\centering
\small
\caption{Executable lifecycle hooks.}
\label{tab:hook-interface}
\begin{tabularx}{\linewidth}{@{}p{0.26\linewidth}X@{}}
\toprule
Hook & Invocation and effect \\
\midrule
\texttt{on\_init} &
Before the first target decision; adds reusable task guidance or a tool hint. \\
\texttt{make\_pre\_hint} &
Before a target decision; injects a state-conditioned message without
executing an action. \\
\texttt{on\_before\_action} &
After the target proposes an action but before environment execution; may
block and reprompt or, where supported, rewrite or force the pending action. \\
\texttt{on\_post\_step} &
After environment feedback; injects recovery guidance or, where supported,
schedules a next action. \\
\bottomrule
\end{tabularx}
\end{table}

The return contract separates contextual guidance from action mediation.
\texttt{make\_pre\_hint} returns a \texttt{message}, and
\texttt{on\_post\_step} may return an \texttt{inject\_hint} effect.
\texttt{block\_and\_prompt} suppresses the current pending action and asks the
frozen target to choose again.  Where enabled, \texttt{rewrite\_action}
replaces that pending action, while \texttt{force\_action} selects a concrete
action for the current or next execution step.  DBBench v1 accepts soft
guidance and blocking but does not execute SQL rewrites or forced commits.

The engineer generates the patch before the rerun and does not participate in
the subsequent task interaction.  It never calls an environment tool or
submits a final task answer directly; the host runtime alone interprets the
hook's structured effects.  Candidates that do not yield an installable
intervention, are behaviorally inert, or do not complete evaluation are
treated as no intervention and receive zero reward.  A valid, complete patch
instead receives its full-batch performance difference, including a negative
value when it causes regressions.

\section{Data Construction and Task Splits}
\label{app:data-splits}

We split benchmark tasks before collecting trajectories or generating harness
patches.  The SFT and online-RL training sets contain the task identities used
to construct their respective training signals.  The validation set is used
only for checkpoint selection, and the test set is used only for final
evaluation.  Table~\ref{tab:task-splits} reports numbers of distinct benchmark
tasks, rather than numbers of trajectories, failure packets, generated
patches, or optimizer samples.

\begin{table}[t]
\centering
\small
\caption{Task-level data splits.  SFT train and RL train are disjoint task
partitions; Train total is their sum.}
\label{tab:task-splits}
\begin{tabular}{lrrrrr}
\toprule
Benchmark & SFT train & RL train & Train total & Validation & Test \\
\midrule
WebShop  & 5,290 & 5,190 & 10,480 & 100 & 500 \\
ALFWorld & 1,380 & 1,280 &  2,660 &  99 & 500 \\
DBBench  & 2,401 & 2,302 &  4,703 & 100 & 300 \\
\midrule
Total    & 9,071 & 8,772 & 17,843 & 299 & 1,300 \\
\bottomrule
\end{tabular}
\end{table}

\paragraph{WebShop.}
We use task indices 0--499 as the fixed test set.  A separate training pool is
randomly divided at the task-batch level into SFT and RL partitions with seed
20260603.  We then reserve 100 tasks from the RL partition for validation with
seed 20260623; the remaining 5,190 tasks form the RL training set.  WebShop
task generation uses goal seed 233 throughout baseline and patched execution.

\paragraph{ALFWorld.}
We stratify by the six ALFWorld task families.  The 500-task test set contains
all 109 tasks from the official \texttt{new\_std} split and a stratified
391-task sample from \texttt{train\_valid}.  With seed 20260614, the remaining
tasks are divided into 1,380 SFT tasks and an RL-side partition; 99 RL-side
tasks are reserved for validation, leaving 1,280 RL training tasks.

\paragraph{DBBench.}
We shuffle the 4,803 available training tasks with seed 20260625 and assign
2,401 tasks to SFT and 2,402 to the RL side.  We reserve 100 RL-side tasks for
validation, leaving 2,302 RL training tasks.  The separate 300-task standard
test set is used only for evaluation.

Teacher filtering, failure-packet construction, benchmark balancing, and
multi-candidate sampling operate within these task partitions.  Their
resulting record counts are therefore training-accounting quantities, not
additional task splits.

\FloatBarrier
\section{Target-Agent Generalization}
\label{app:cross-target}

We apply the single learned editing policy to a broad set of target agents that
are never used during training.  For every target, the engineer reads that
target's own failure traces and generates target-specific patches; this
measures transfer of the editing \emph{policy}, not reuse of one fixed patch.
Table~\ref{tab:supp-cross-target} reports the per-target, per-benchmark task
success before and after installing the target-specific patches, providing the
absolute levels behind the delta heatmap in the main paper.  WebShop uses the
fixed-seed 500-task rerun (goal seed 233); ALFWorld and DBBench use their
respective test sets.  The \(\boldsymbol{\Delta}\)\,\textbf{Avg.} column is the
equal-weight average of the three benchmark deltas, and \(\dagger\) marks the
primary Qwen3.5-9B target that is also used in the main results.

\begin{table}[t]
\centering
\small
\setlength{\tabcolsep}{5.5pt}
\caption{Target-agent generalization: per-benchmark task success rate before and
after installing target-specific patches (\%), with the equal-weight benchmark
average of the deltas (\(\Delta\)\,Avg., pp).  Rows are grouped by model family;
\(\dagger\) marks the primary Qwen3.5-9B target.  WebShop uses the fixed-seed
500-task rerun; ALFWorld and DBBench use their test sets.}
\label{tab:supp-cross-target}
\begin{tabular}{@{}lccccccr@{}}
\toprule
& \multicolumn{2}{c}{\textbf{WebShop}} & \multicolumn{2}{c}{\textbf{ALFWorld}} & \multicolumn{2}{c}{\textbf{DBBench}} & \\
\cmidrule(lr){2-3}\cmidrule(lr){4-5}\cmidrule(lr){6-7}
\textbf{Target agent} & Before & After & Before & After & Before & After & \(\boldsymbol{\Delta}\)\,\textbf{Avg.} \\
\midrule
Llama-3.1-8B & 21.2 & 31.0 & 2.0 & 10.0 & 16.7 & 30.3 & +10.5 \\
Llama-3.1-70B & 39.2 & 38.8 & 21.0 & 35.2 & 31.7 & 33.7 & +5.3 \\
Llama-3.2-1B & 0.0 & 0.0 & 0.0 & 0.2 & 8.0 & 14.3 & +2.2 \\
Llama-3.2-3B & 8.4 & 13.6 & 1.2 & 7.2 & 7.0 & 16.7 & +7.0 \\
Llama-3.3-70B & 35.4 & 41.8 & 27.4 & 46.2 & 60.3 & 63.0 & +9.3 \\
\midrule
Gemma-3-1B & 0.0 & 0.0 & 0.0 & 1.2 & 0.3 & 2.7 & +1.2 \\
Gemma-3-4B & 13.8 & 28.4 & 4.0 & 10.4 & 14.3 & 29.3 & +12.0 \\
Gemma-3-12B & 22.8 & 32.2 & 12.8 & 18.8 & 41.0 & 56.7 & +10.4 \\
Gemma-3-27B & 27.8 & 37.2 & 18.6 & 29.4 & 52.3 & 60.0 & +9.3 \\
Gemma-4-12B-it & 39.4 & 39.4 & 35.8 & 55.2 & 61.3 & 67.7 & +8.6 \\
Gemma-4-26B-A4B-it & 39.2 & 39.6 & 49.2 & 65.4 & 60.0 & 69.0 & +8.5 \\
Gemma-4-31B-it & 42.0 & 41.8 & 54.4 & 73.2 & 65.7 & 69.0 & +7.3 \\
\midrule
Qwen2.5-72B & 37.8 & 39.6 & 70.8 & 68.8 & 51.3 & 54.7 & +1.0 \\
Qwen3-4B & 25.0 & 38.6 & 22.8 & 29.1 & 38.7 & 55.7 & +12.3 \\
Qwen3-8B & 30.6 & 34.6 & 23.0 & 27.7 & 49.3 & 57.3 & +5.6 \\
Qwen3-14B & 33.8 & 37.2 & 22.3 & 39.0 & 50.0 & 60.0 & +10.0 \\
Qwen3.5-4B & 33.2 & 36.8 & 20.7 & 38.6 & 60.7 & 65.0 & +8.6 \\
\textbf{Qwen3.5-9B}$^{\dagger}$ & 31.2 & 42.2 & 40.6 & 53.2 & 61.0 & 65.3 & +9.3 \\
Qwen3.5-27B & 42.0 & 42.0 & 72.4 & 81.3 & 70.3 & 72.3 & +3.6 \\
Qwen3.5-35B-A3B & 30.6 & 31.8 & 62.2 & 69.2 & 62.7 & 68.7 & +4.7 \\
Qwen3.6-27B & 43.4 & 44.2 & 70.6 & 78.6 & 69.7 & 72.7 & +3.9 \\
\midrule
\textbf{Mean, 20 unseen targets} & 28.3 & 32.4 & 29.4 & 39.1 & 43.6 & 50.9 & \textbf{+7.1} \\
\textbf{Mean, all 21 targets} & 28.4 & 32.9 & 30.0 & 39.8 & 44.4 & 51.6 & \textbf{+7.2} \\
\bottomrule
\end{tabular}
\end{table}

Every target-level average is positive, and across the full \(21\times3\) matrix
\(56\) of \(63\) target--benchmark combinations improve, four are unchanged
(all on WebShop), and the three small regressions are WebShop on Llama-3.1-70B
(\(-0.4\)), ALFWorld on Qwen2.5-72B (\(-2.0\)), and WebShop on Gemma-4-31B-it
(\(-0.2\)).  The benchmark-averaged gain across the twenty unseen targets is
\(7.06\) points, showing that a single training recipe transfers across model
families and scales without any per-target retuning.

\section{Held-Out Task Generalization}
\label{app:heldout}

We further test whether patches inferred from a handful of failures improve
\emph{unseen tasks}.  For each benchmark and seed, every engineer observes the
same ten failures from the frozen Qwen3.5-9B target, generates one
benchmark-specific patch, and applies it to all remaining tasks; we repeat the
protocol over three matched evidence seeds.  Invalid patches are counted as no
intervention (zero delta).  Table~\ref{tab:supp-heldout} reports the pooled
held-out change (1,270 tasks) and, for reference, the full-split change
including the ten evidence tasks (1,300 tasks); the error term is the sample
standard deviation across the three seeds.

\begin{table}[t]
\centering
\small
\setlength{\tabcolsep}{4.5pt}
\caption{Held-out-task generalization from sparse failure evidence
(\(\Delta\) success, pp; mean \(\pm\) sample std over three seeds).  Valid counts
the seed-benchmark patches that installed a real intervention out of nine.}
\label{tab:supp-heldout}
\begin{tabular}{@{}lccc@{}}
\toprule
\textbf{Engineer} & \textbf{Valid} & \textbf{Held-out (1,270)} & \textbf{Full split (1,300)} \\
\midrule
\textbf{Harness-R1} & 9/9 & \textbf{+8.9$\pm$1.5} & \textbf{+9.2$\pm$1.5} \\
Qwen3.5-397B & 8/9 & $-4.3\pm2.5$ & $-3.9\pm2.5$ \\
DeepSeek-V4-Pro & 6/9 & $-0.4\pm3.6$ & $-0.2\pm3.5$ \\
\bottomrule
\end{tabular}
\end{table}

Harness-R1 improves pooled held-out success by \(8.9\pm1.5\) points and is
positive on every seed, whereas both frontier engineers average negative and
straddle zero across seeds.  The larger frontier spreads (\(\pm2.5\) and
\(\pm3.6\)) reflect swings between marginal gains and sizable regressions, so
converting sparse failures into a broadly useful edit is a capability that scale
alone does not confer.

\section{Lifecycle-Position Ablation}
\label{app:lifecycle}

To attribute the improvement to specific intervention points, we hold the frozen
vanilla target and the generated patches fixed and disable one lifecycle
position at a time, alongside a no-intervention control and the full patch.
Each configuration reruns the target three times per benchmark, and we report the
equal-benchmark-weighted success rate used in the main results.
Table~\ref{tab:supp-lifecycle} lists the per-benchmark and averaged success
behind the ablation figure in the main paper.

\begin{table}[t]
\centering
\small
\setlength{\tabcolsep}{3.2pt}
\caption{Fixed-patch lifecycle-position ablation on the vanilla target
(success rate, \%).  The \textbf{Avg.} column is the equal-weight benchmark
average; parenthesized values are the drop relative to the full patch.}
\label{tab:supp-lifecycle}
\begin{tabular}{@{}lrrrr@{}}
\toprule
\textbf{Configuration} & \textbf{WebShop} & \textbf{ALFWorld} & \textbf{DBBench} & \textbf{Avg.} \\
\midrule
\textbf{Full patch} & 41.6 & 52.1 & 65.4 & \textbf{53.1} \\
w/o episode init & 41.5 & 51.3 & 63.8 & 52.2 ($-$0.9) \\
w/o pre-decision & 41.5 & 50.4 & 65.6 & 52.5 ($-$0.6) \\
w/o pre-action & 31.5 & 50.7 & 65.4 & 49.2 ($-$3.9) \\
w/o post-feedback & 41.7 & 41.9 & 65.8 & 49.8 ($-$3.3) \\
\midrule
No intervention & 31.8 & 40.7 & 60.1 & 44.2 \\
\bottomrule
\end{tabular}
\end{table}

The full patch reaches \(53.1\%\) average success, \(8.9\) points above no
intervention.  Removing pre-action mediation or post-feedback recovery causes
the largest drops (\(3.9\) and \(3.3\) points), while removing episode
initialization or pre-decision costs only \(0.9\) and \(0.6\) points.  The
dominant position is environment-dependent: pre-action mediation matters most on
WebShop (success falls from \(41.6\) to \(31.5\)), whereas post-feedback
recovery matters most on ALFWorld (\(52.1\) to \(41.9\)).  Because a single patch
can coordinate several positions, these effects are conditional and should not
be summed into a universal importance ranking.

\FloatBarrier
\section{Qualitative Case Studies}
\label{app:cases}

We examine three stored evaluations from the validation-selected Harness-R1
engineer used for the main frozen-target results.  In every case, the target
agent is the same frozen Qwen3.5-9B model before and after patch installation,
and the patched run uses the same ten tasks as its baseline evidence.  We
inspect the runtime trace in addition to the generated code, so that an
intended edit is not mistaken for an intervention that actually executed.
These cases illustrate distinct mechanisms and limitations; aggregate claims
are based on the full evaluations in the main paper rather than on the selected
examples.

\begin{table}[t]
\caption{Overview of the qualitative cases.}
\label{tab:case-overview}
\centering
\small
\begin{tabularx}{\linewidth}{@{}lcccX@{}}
\toprule
Environment & Success (before) & Success (after) & \(\Delta\) & Primary behavior illustrated \\
\midrule
WebShop  & 2/10 & 5/10 & +3 & A narrow guard delays purchase until required options are selected. \\
ALFWorld & 1/10 & 6/10 & +5 & State tracking, stage-specific guidance, and an action guard form a closed loop. \\
DBBench  & 4/10 & 6/10 & +2 & Schema recovery and format-preserving mutation outperform a valid GLM-5.2 edit. \\
\bottomrule
\end{tabularx}
\end{table}

\subsection{WebShop: Correcting a Premature Purchase}

\paragraph{Observed failure and generated edit.}
In WebShop batch 008, several trajectories reached a relevant, in-budget
product but issued \texttt{Buy Now} before choosing an option required by the
instruction.  The engineer generated a single pre-action intervention.  It
activates only for a normalized \texttt{Buy Now} action and blocks the action
when either the current price exceeds the budget or a required product option
remains unselected.  The resulting message asks the target to choose a visible
matching option, or return to search if no such option exists.

\paragraph{Task-level behavior.}
One task requests a synthetic hairpiece with a black-brown color and a price
below \$40.  The baseline target finds a suitable product but purchases it
without selecting the color, receiving a partial reward of 0.667.  With the
patch installed, the target initially proposes the same premature purchase.
The guard blocks it, after which the target selects \texttt{black brown} and
then purchases the item, receiving reward 1.0.

\begin{center}
\small
\begin{tabularx}{0.92\linewidth}{@{}p{0.16\linewidth}Xr@{}}
\toprule
Run & Relevant action sequence & Reward \\
\midrule
No intervention &
Search $\rightarrow$ open product $\rightarrow$ \texttt{Buy Now} with color unselected &
0.667 \\
Harness-R1 &
Search $\rightarrow$ open product $\rightarrow$ attempted \texttt{Buy Now}
$\rightarrow$ guard message $\rightarrow$ select \texttt{black brown}
$\rightarrow$ \texttt{Buy Now} &
1.000 \\
\bottomrule
\end{tabularx}
\end{center}

Across the ten-task batch, the same guard raises full successes from 2 to 5
and mean WebShop reward from 0.682 to 0.768, while preserving both tasks that
were already fully successful.  This case shows that an effective harness edit
need not replace the target's policy with a large controller: a low-bandwidth
intervention at the point of an unsafe action can preserve the target's search
behavior while changing the final outcome.

\subsection{ALFWorld: Coordinating Multiple Lifecycle Positions}

\paragraph{Observed failure and generated edit.}
ALFWorld batch 045 contains recurrent failures in which the target finds an
object but omits a required transformation, moves toward the wrong receptacle,
or enters a transform--place loop.  The generated patch coordinates four
positions in the execution lifecycle:

\begin{center}
\small
\begin{tabularx}{0.92\linewidth}{@{}p{0.27\linewidth}X@{}}
\toprule
Lifecycle position & Installed behavior \\
\midrule
Episode initialization &
Initialize the current stage and provide the reusable
find--take--transform--place ordering. \\
After environment feedback &
Update whether the target is held, the required transformation is complete,
and the destination has been reached. \\
Before model decision &
Inject a stage-specific hint for the next unresolved subgoal. \\
Before environment execution &
Block a premature placement or a placement at the wrong destination. \\
\bottomrule
\end{tabularx}
\end{center}

\paragraph{Runtime behavior.}
The intervention trace records 56 stage hints or guard messages across the
batch.  For a task requiring a hot mug in a cabinet, the target successively
receives guidance to take the mug, heat it using the microwave, navigate to
the cabinet, and place the mug.  For a task requiring a cooled egg on a
countertop, the target attempts to place the egg on a dining table; the
pre-action guard rejects that destination, and the subsequent trajectory
places the egg on the requested countertop.

\begin{center}
\small
\begin{tabularx}{0.92\linewidth}{@{}p{0.21\linewidth}Xc@{}}
\toprule
Example & Effective intervention sequence & Outcome \\
\midrule
Hot mug in cabinet &
Take-target hint $\rightarrow$ heat-at-microwave hint $\rightarrow$
go-to-destination hint $\rightarrow$ put-target hint &
Failure $\rightarrow$ success \\
Cooled egg on countertop &
Stage hint $\rightarrow$ attempted wrong placement $\rightarrow$
destination guard $\rightarrow$ corrected placement &
Failure $\rightarrow$ success \\
\bottomrule
\end{tabularx}
\end{center}

At batch level, the patch rescues six baseline failures but regresses one
baseline success, producing a net change from 1/10 to 6/10.  It also fails to
resolve every task: one two-object trajectory continues to alternate between
destination and placement guidance.  The example therefore demonstrates a
genuine closed-loop harness policy, while also showing that stage tracking can
remain imperfect.

\subsection{A Failure Case of Direct Harness Editing}

An off-the-shelf model can access the complete lifecycle interface yet still
produce harmful interventions.  On ALFWorld, Gemini-3.5-Flash receives full
failure evidence and may edit all four lifecycle positions.  Its patches reduce
success from 208/500 (41.6\%) to 177/500 (35.4\%), a drop of 6.2 percentage
points.  Among 39 valid patches, 21 reduce batch success, 12 improve it, and 6
leave it unchanged.

\paragraph{Overgeneralized action intervention.}
The largest regression occurs in a batch whose success falls from 7/10 to
0/10.  The patch installs broad \texttt{on\_before\_action} rules that force
actions from a locally plausible stage estimate.  On two-object tasks, it
prematurely places the first object rather than collecting both objects before
placement, overriding decisions that the frozen target agent previously
executed correctly.  This case illustrates why execution traces and a powerful
base model alone do not yield a reliable harness editor: a plausible diagnosis
can still compile into overly aggressive runtime behavior.  Harness-R1 instead
post-trains the editing policy on realized task outcomes, directly penalizing
patches that degrade rerun performance.

\subsection{DBBench: Preserving Schema and Stored-Value Conventions}

\paragraph{Observed failure and generated edit.}
DBBench batch 022 contains recurring failures around multi-word identifiers,
schema recovery, and exact mutation values.  Harness-R1 generates a
stage-aware patch that recommends schema inspection after identifier errors,
asks the target to inspect the affected row before mutation, and verifies the
stored value before commit.  The patch raises the frozen target from 4/10 to
6/10.  On the same baseline evidence and tasks, a valid GLM-5.2 patch raises
the result only to 5/10.

\paragraph{Task-level contrast.}
One task asks the agent to change the length of the
\texttt{Moosehead Grand Prix} entry in the multi-word table
\texttt{Race Schedule}.  The no-intervention trajectory recovers the quoted
table name and observes the existing value \texttt{3 Hours}, but writes
\texttt{4 hours}; the exact-format evaluator marks the task incorrect.
GLM-5.2 supplies general backtick and mutation-verification guidance, yet its
guided trajectory makes the same lower-case write.  Harness-R1 first triggers
schema recovery, inspects the existing row, writes \texttt{4 Hours} to match
the stored convention, and verifies the row before committing.

\begin{center}
\small
\begin{tabularx}{0.88\linewidth}{@{}Xcc@{}}
\toprule
Runtime condition & Batch success & Example outcome \\
\midrule
No intervention & 4/10 & \texttt{4 hours} (failure) \\
GLM-5.2 patch & 5/10 & \texttt{4 hours} (failure) \\
Harness-R1 patch & 6/10 & \texttt{4 Hours} (success) \\
\bottomrule
\end{tabularx}
\end{center}

This paired example does not rely on an invalid competitor output: both
engineers produce executable patches.  The difference is that the
Harness-R1-guided run converts schema and row evidence into the exact stored
representation required by the task.

\subsection{Cross-Case Interpretation}

\paragraph{WebShop.}
The recurring failure is premature purchase.  Harness-R1 installs a narrow
action guard conditioned on runtime predicates, although the guard cannot
repair an earlier choice of the wrong product.

\paragraph{ALFWorld.}
The recurring failures are omitted transformations and incorrect placement.
Harness-R1 combines persistent stage state, targeted hints, and a placement
guard, while routing and two-object state can still cause regressions or
loops.

\paragraph{DBBench.}
The recurring failures involve multi-word identifiers and exact mutation
formats.  Harness-R1 uses schema recovery, row inspection, and
format-preserving verification; some value conventions still require
stronger neighborhood-level checks.

Together, the cases show that Harness-R1 learns environment-dependent editing
policies rather than one fixed prompt.  Outcome-grounded post-training
increases useful executable edits without guaranteeing complete or
regression-free rules.

\end{document}